\def\year{2021}\relax
\documentclass[letterpaper]{article} 
\usepackage{aaai21}  
\usepackage{times}  
\usepackage{helvet} 
\usepackage{courier}  
\usepackage[hyphens]{url}  
\usepackage{graphicx} 
\usepackage{natbib}  
\usepackage{caption} 
\usepackage{subcaption}
\usepackage[table]{xcolor}
\usepackage{paralist}
\usepackage{pifont}
\newcommand{\cmark}{\ding{51}}%
\newcommand{\xmark}{\ding{55}}%

\newcommand{\doubt}[1]{\textcolor{red}{DOUBT: #1}}
\newcommand{\blu}[1]{\textcolor{black}{#1}}
\newcommand{\purpl}[1]{\textcolor{black}{#1}}

\title{Two-Face: Adversarial Audit of Commercial Face Recognition Systems}
\author{
    Siddharth D Jaiswal,
    Karthikeya Duggirala,
    Abhisek Dash,
    Animesh Mukherjee
    \\
}
\affiliations{
    Indian Institute of Technology Kharagpur, India
}

\begin{document}

\maketitle

\begin{abstract}
Computer vision applications like automated face detection are used for a variety of purposes ranging from unlocking smart devices to tracking potential persons of interest for surveillance. Audits of these applications have revealed that they tend to be biased against minority groups which result in unfair and concerning societal and political outcomes. Despite multiple studies over time, these biases have not been mitigated completely and have in fact increased for certain tasks like age prediction. While such systems are audited over benchmark datasets, it becomes necessary to evaluate their robustness 
for adversarial inputs. In this work, we perform an extensive adversarial audit on multiple systems and datasets, making a number of concerning observations -- there has been a drop in accuracy for some tasks on \textsc{CelebSET} dataset since a previous audit. While there still exists a bias in accuracy against individuals from minority groups for multiple datasets, a more worrying observation is that these biases tend to get exorbitantly pronounced with adversarial inputs toward the minority group. We conclude with a discussion on the broader societal impacts in light of these observations and a few suggestions on how to collectively deal with this issue.\footnote{\textcolor{red}{This work has been accepted for publication at ICWSM 2022}}


\end{abstract}

\noindent

\section{Introduction}
\label{sec:intro}
AI based systems are increasingly being used to assist human decision makers in hiring, granting loans, recidivism prediction; i.e., decisions that have larger consequences on the life and livelihood of the stakeholders~\cite{o2016weapons,noble2018algorithms}.
To this end, even systems that are not specifically geared for such high stakes decision making sometimes end up being a part of the decision pipeline. For example, while Facial Recognition Systems (\textbf{FRSs}) are not meant for identifying individuals in the criminal justice system, it is very likely that such systems are being used to identify suspects in reality
~\cite{Sangomla2020Big}. Thus, an error in the outcome of a face recognition algorithm 
can have serious repercussions. For example, an innocent person can be wrongfully accused of crime or a criminal might be set free due to their inaccurate identification. 

\noindent
\textbf{Societal concerns surrounding usage of FRSs}: FRSs are widely deployed for tasks such as: estimating customer satisfaction, demographic characteristics of population, tracking individuals etc.~\cite{nagpal2019deep}. These are done under the broader purview of customer understanding, research 
and security scenarios~\cite{Amazon2021FAQ,MS2021What,FPP2021What}.  
However, the reality of FRS deployments, reveals that often their usages are vulnerable to abuse, causing intentional or unintentional discrimination against
marginalized groups~\cite{benjamin2019race,buolamwini2018gender, raji2020saving}. 
In fact in some cases, FRSs with accuracy lower than 1\% have been used to differentiate between conscientious citizens and anarchic rioters in the context of civil riots~\cite{Sangomla2020Big}. 
The stakes are particularly high considering Amazon, HireVue etc., are also selling their services to help in policing and hiring candidates~\cite{Buolamwini2018What,Snow2018Amazon}. 

\noindent
\textbf{Prior audits of FRSs}: Like most AI based systems, these facial recognition systems are trained on large scale data. However, multiple recent studies have repeatedly shown how such systems are prone to be biased against certain sections of the population~\cite{mehrabi2019survey,bolukbasi2016man}. Such biases can be attributed to either biased training instances or procedural flaw in the underlying algorithms. The lack of open access to such data driven algorithms or any established model cards~\cite{mitchell2019model} further impede the public know-how of the working of these systems and source of these biases~\cite{sandvig2014auditing}.
Thus, the only way to evaluate the bias in these software is to conduct black box audits~\cite{sandvig2014auditing,dash2019network,dash2021when}. To this end, prior audits have shown commercial face detection systems by IBM, Microsoft, Amazon, and Face++ to be biased to faces belonging to non-White ethnic groups and / or female gender~\cite{buolamwini2018gender,raji2020saving}.

\noindent
\textbf{Potential of temporal and emergent biases}: While these systems have been previously audited by researchers, the extent of biases and even performance of the systems seem to vary with time. Further, in the aim of improving on one aspect (e.g., fairness in gender detection), systems can potentially overlook several other aspects such as: performance on other tasks (e.g., age prediction, emotion detection) and / or are prone to exacerbate reverse discrimination.
As per the taxonomy of biases proposed by~\cite{mehrabi2019survey}, these concerns fall in the scope of emergent bias. Thus, we posit systems of such significant societal consequences should be periodically audited for their performance and biases. This brings us to our first research question.

\noindent\textbf{RQ(1)}: \textit{Do the performances (and biases (if any)) of commercial FRSs vary over time (for the same dataset)}?

\noindent
\textbf{Need for adversarial audits}: Note that the images taken by surveillance cameras are often prone to different kind of noises caused due to several practical reasons. For example, their lenses are exposed to environmental elements, e.g., rain, dirt, storms etc.; further they do not cover all areas of the locality with the same resolution. Hence, auditing the performance on well formed, noiseless images is not enough. We therefore posit that their performance on \textit{noisy or perturbed} images is of equal importance. This provides the context for our second research question.

\noindent\textbf{RQ(2)}: \textit{Are the commercial FRSs immune to perturbation or noises induced in captured images}?

To this end, adversarial attack on images has been extensively used in the context of adversarial learning in AI/ML~\cite{vakhshiteh2020adversarial}. Taking a leaf out of such techniques, we also audit the commercial FRSs for their performance and biases (if any) by poisoning images using adversarial noise models (more details in the section on datasets and adversarial inputs).

\noindent
\textbf{Current work}: Keeping the aforementioned considerations in mind, in this work, we perform an audit study of three such commercial facial detection systems by Microsoft, Amazon and Face++. We demonstrate our experiments on publicly available \textsc{CelebSET}~\cite{raji2020saving}, \blu{\textsc{FairFace}~\cite{sixta2020fairface}} and \textsc{Chicago Face Database} (CFD)~\cite{ma2015chicago,ma2020chicago,lakshmi2021india} datasets. 
\blu{Note that our work focuses on face identification task by FRSs. Additional objectives of person / face re-identification or matching is beyond the scope of the current work. We leave these additional objectives and audits thereof for an immediate future work.}
We summarize our contributions and observations as follows.


\begin{compactitem}
	\item By auditing the \textsc{CelebSET} dataset, we demonstrate the inconsistent performance of FRSs as compared to the prior audit~\cite{raji2020saving}. While we observe marginal drop in performance in the gender classification task, the drop in performance for age detection task is alarming ($\ge 15\%$ drop for both Amazon, and Microsoft).
	
	\item By generating noisy adversarial images with the help of easily understandable open source software \textsc{GIMP}, we conduct audits on \textsc{CelebSET}, \blu{\textsc{FairFace}} and \textsc{CFD}.
	\begin{compactitem}
		\item \textsc{CelebSET}: We observe a significant drop in accuracy for all tasks, with the three FRSs performing particularly poorly on RGB noise. This raises serious questions on the robustness of such systems.
		\item \blu{\textsc{FairFace}: We observe a significant drop in accuracy on all but one tasks for the images poisoned with RGB noise for all FRSs. The FRSs perform slightly better on the spread noise. Thus the accuracy of the FRSs is highly dependent upon the type of noise.}
		\item \textsc{CFD}: The FRSs are audited for two tasks and exhibit a slightly more robust behaviour with a significant drop in accuracy only for Newsprint noise. Thus the accuracy of the FRSs is dependent on a combination of the dataset and noise.
	\end{compactitem}
	
	\item Further involved analyses of the predictions suggest that the disparity in intersectional accuracy for minority groups increase drastically with the adversarial inputs. In \textsc{CelebSET}, Microsoft FRS reports a disparity of $\geq 30\%$ for all tasks between people belonging to `White' ethnicity and `Black' ethnicity. \blu{In \textsc{FairFace}, the disparity is always against people of color. For Microsoft FRS, the disparity is always against `Black' males while the other FRSs display disparity against other ethnic groups as well. \purpl{For instance, gender detection on the AWS FRS exhibits a disparity as high as $74\%$ against `East Asian' males.}}
	Similarly, in \textsc{CFD}, the disparity in gender accuracy is $93\%$ against `Black' females and $44\%$ for age prediction against `Black' males for Face++ FRS. This shows that adversarial noises also widen the disparity against minority groups along with low prediction accuracy.
\end{compactitem}

To the best of our knowledge, this is the first attempt to perform an adversarial audit of the commercial facial recognition systems. We believe that observations in this paper might ignite further discussion and deliberation among practitioners and researchers regarding the ethical and governance aspect of such high-stakes technology.

\section{Background and related work}
\label{sec:rel-work}

\begin{table}[t]
	\noindent
	\small
	\centering
	\begin{tabular}{ |c|c|c|c|}
		\hline
		{\bf Tasks} & {\bf Microsoft } &  {\bf Amazon } &  {\bf Face++}  \\
		\hline
		Gender detection & \cmark & \cmark & \cmark \\
		\hline
		Age detection & \cmark & \cmark & \cmark \\
		\hline
		Smile detection & \cmark & \cmark & \cmark \\
		\hline
		Emotion detection & \cmark & \cmark & \cmark \\
		\hline
		Bounding box detection & \cmark & \cmark & \cmark \\
		\hline
		Facial landmarks & \cmark & \cmark & \cmark \\
		\hline
		Beauty score & \xmark & \xmark & \cmark \\
		\hline
		Ethnicity & \xmark & \xmark & \cmark \\
		\hline
		Skin status & \xmark & \xmark & \cmark \\
		\hline
	\end{tabular}	
	\caption{{\bf Some of the important tasks performed by each of the individual commercial Facial Recognition Systems. We compare them on the first three tasks keeping our audit in-line with the prior audits. }}
	\label{Tab: Tasks-FRS}
	\vspace{-5 mm}
\end{table}

In this section, we first provide a brief background about the commercial FRSs that we study. This is followed by a review of prior literature on two related strands of research -- (a)~bias in FRSs, (b)~adversarial attacks in computer vision. 
\subsection{Facial recognition systems (FRSs)}
In this work we specifically study the following FRSs 
\begin{compactitem}
	\item Amazon AWS Rekognition~\cite{aws_rekognition}
	\item Microsoft Azure Face~\cite{microsoft_face}
	\item Face++ Detect~\cite{facepp}
\end{compactitem}
These systems are readily available for individual and commercial users. 
They provide a variety of services like identification of gender, age, emotion, facial features, and even beauty~\cite{arwa_guardian2021,tate_mit2021} 
on input images. 

The list of different prominent tasks performed by each of these facial recognition system is listed in Table~\ref{Tab: Tasks-FRS}. Even though the systems offer a number of tasks where they are comparable, keeping in line with the prior research by ~\cite{buolamwini2018gender, raji2020saving}, we compare them primarily on the first three tasks, i.e., (1)~gender detection, (2)~age detection, and (3)~smile detection 
\footnote{In CFD dataset, as all images have neutral face expression, we have not performed comparison on smile detection task.}.
Note that even though there are some potentially controversial tasks mentioned in Table~\ref{Tab: Tasks-FRS} (last three rows), since they are not performed by two of the FRSs, we exclude them from our further investigation in this work. However, we acknowledge that these tasks need to be studied and investigated on their own as part of future research in this domain.

\noindent
\textbf{Difference in outputs for age prediction: }For the age prediction task while Microsoft and Face++ systems report an exact predicted age, Amazon system reports a window which ranges between 10-18 years. For example, if the returned window is $[x, x+10]$, the system predicts the age of the subject to be at least $x$ years and at most $x + 10$ years.
In this study, we use the median value from Amazon system's predicted range as the age of the subject in the image for comparison with the other two systems.


\subsection{Bias in FRSs}

AI based systems are data driven and are shown to be prone to inadvertent consequences such as bias and unfairness. 
FRSs are no exception. Even though these systems are reported to achieve an overall accuracy upwards of 80\%, the problem lies underneath. 
\citet{buolamwini2018gender} showed the disparity of these systems for the first time by highlighting the difference in performance for the dark-skinned population as compared to the fair-skinned counterparts. There have been further follow up studies which have audited the same and other commercial FRSs \cite{JungICWSM2018,raji2019action,kyriakou2019,raji2020saving,sixta2020fairface} which showed even after the first audit, these disparities prevailed and also reported different performance diagnostics at different points of time. 

On the positive impact of such studies, IBM has stopped developing facial recognition technology~\cite{peters_verge2020}
and there have been calls by watchdog organizations for others to follow suit~\cite{stop_2021}.

\noindent
\blu{\textbf{Relevance to the ICWSM community: }Computational social science community has shown continuous interest in this area of research. For instance, recent works
~\cite{KyriakouICWSM2019,BarlasICWSM2019} have explored concerns over fairness in image tagging applications by commercial FRSs. On the other hand, many studies have used these FRSs, especially \textsc{Face++}, as a part of their research pipeline for analysing the different attributes of Twitter~\cite{VikatosHT2017,MessiasWI2017} and Instagram users~\cite{PangBigData2015}. We posit that a biased FRS can lead to researchers making false conclusions.}

Although most studies highlighted above are directed toward the first research question we posed, the variation in their findings further highlight the importance of regular third-party audits of these commercial systems. 

\subsection{Adversarial attack in computer vision}
Adversarial attacks have been leveraged to test the robustness of many computer vision based systems \cite{vakhshiteh2020adversarial}. \blu{In these attacks, some noise is intentionally added to the images~\cite{goodfellow2014explaining,xiao2019meshadv,qiu2019semanticadv,Xu_2020_CVPR,maesumi2021learning,Duan_2020_CVPR} or videos~\cite{jiang2019black,chen2021appending} and then they are fed to the system to check how immune or robust they are to these changes.}
These may also be used to increase privacy for non-consenting individuals~\cite{equalais_2021}.
These techniques may range from being as simple as directly modifying the image file's bits, using image manipulation software like \textsc{Gimp}~\cite{gimp}, using libraries like \cite{bloice2019augmentor}, 
or as sophisticated and involved as using game-theoretic techniques \cite{oh2017adversarial}, and deep learning techniques \cite{chandrasekaran2020faceoff,goel2018smartbox,garofalo2018fishy,bose2018adversarial,massoli2021detection,xiao2019meshadv,jiang2019black,qiu2019semanticadv,Xu_2020_CVPR,maesumi2021learning,Duan_2020_CVPR,chen2021appending}.
\blu{We do not explore sophisticated adversarial attacks as part of this work. Instead we focus on simple attacks that may be a result of real world situations, as discussed later.}

As part of our second research question, we use some rudimentary noise models using \textsc{Gimp} (discussed further in the dataset section) to perturb the input facial images and see how robust the commercial FRSs are to such noises. We not only evaluate the overall robustness of the systems but also execute involved analyses regarding the disparity of performance among intersectional groups. We call this audit -- `\textbf{adversarial audit}' and to the best of our knowledge, this is the first attempt to understand the robustness of these FRSs under an adversarial audit. 


\section{Datasets and adversarial inputs}
\label{sec:dataset-perturb}
\begin{figure}[t]
	\centering
	\begin{subfigure}{0.32\columnwidth}
		\includegraphics[width= \textwidth, height=3cm]{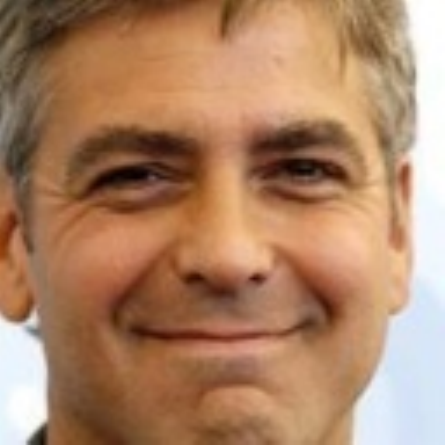}
		\vspace*{-4mm}
		\caption{}
		\label{fig1:a}
	\end{subfigure}%
	~\begin{subfigure}{0.32\columnwidth}
		\centering
		\includegraphics[width= \textwidth, height=3cm]{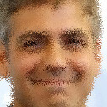}
		\caption{}
		\label{fig1:b}
	\end{subfigure}
	~\begin{subfigure}{0.32\columnwidth}
		\centering
		\includegraphics[width= \textwidth, height=3cm]{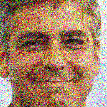}
		\caption{}
		\label{fig1:c}
	\end{subfigure}

	\begin{subfigure}{0.32\columnwidth}
		\includegraphics[width= \textwidth, height=3cm]{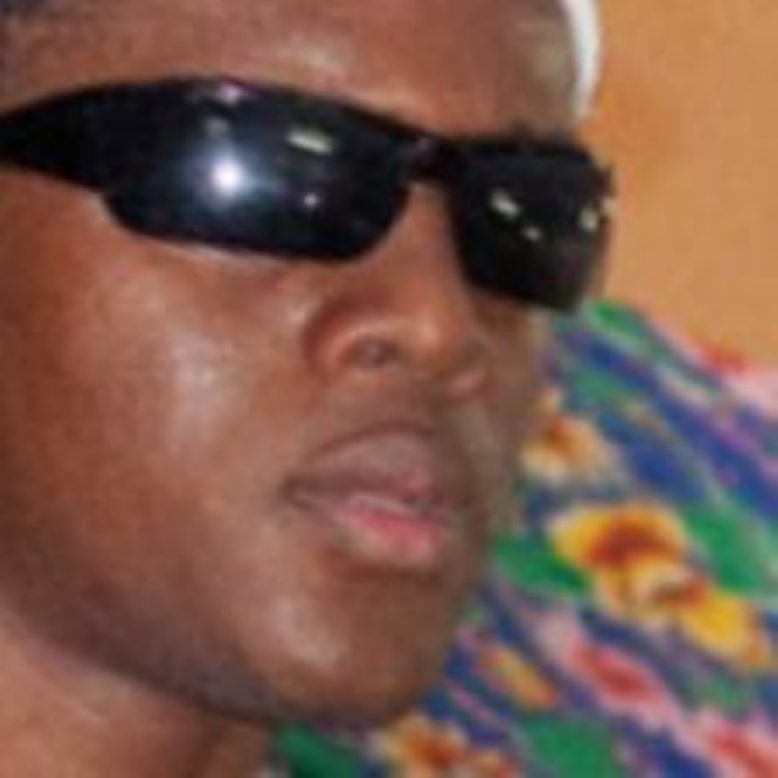}
		\vspace*{-4mm}
		\caption{}
		\label{fig1:d}
	\end{subfigure}%
	~\begin{subfigure}{0.32\columnwidth}
		\centering
		\includegraphics[width= \textwidth, height=3cm]{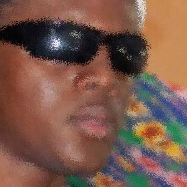}
		\caption{}
		\label{fig1:e}
	\end{subfigure}
	~\begin{subfigure}{0.32\columnwidth}
		\centering
		\includegraphics[width= \textwidth, height=3cm]{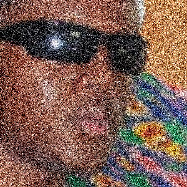}
		\caption{}
		\label{fig1:f}
	\end{subfigure}
	
	\begin{subfigure}{0.32\columnwidth}
		\includegraphics[width= \textwidth, height=3cm]{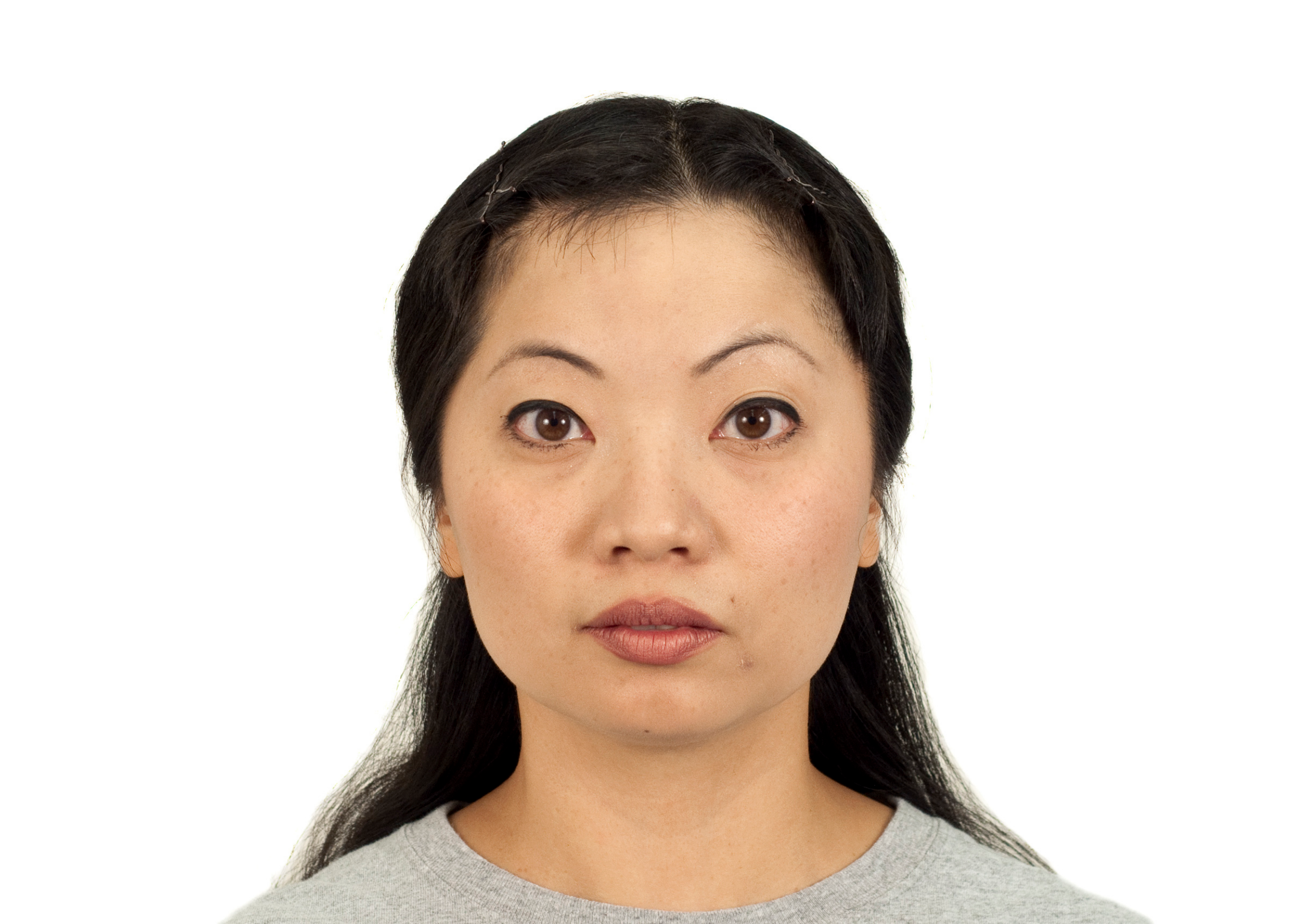}
		\vspace*{-4mm}
		\caption{}
		\label{fig1:d}
	\end{subfigure}%
	~\begin{subfigure}{0.32\columnwidth}
		\centering
		\includegraphics[width= \textwidth, height=3cm]{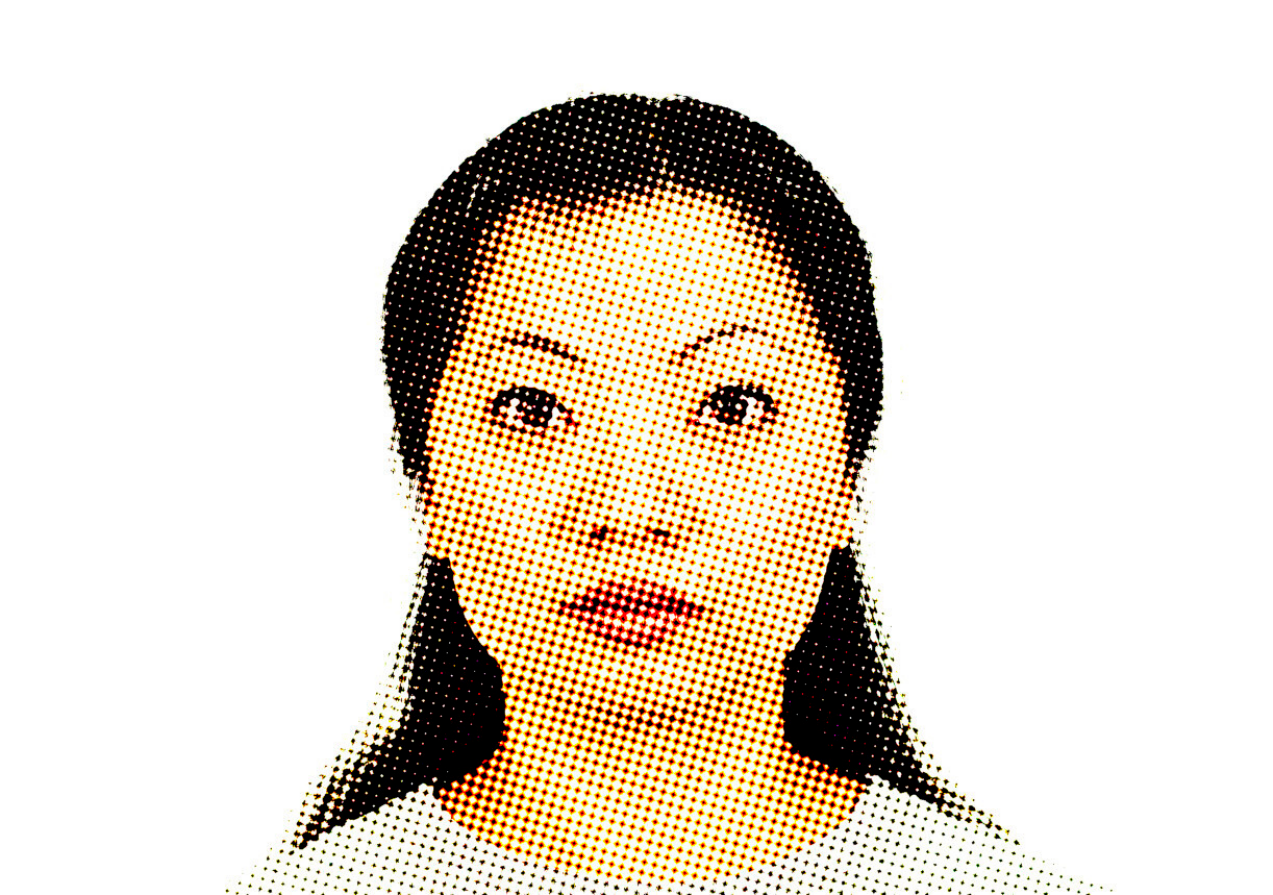}
		\caption{}
		\label{fig1:e}
	\end{subfigure}
	~\begin{subfigure}{0.32\columnwidth}
		\centering
		\includegraphics[width= \textwidth, height=3cm]{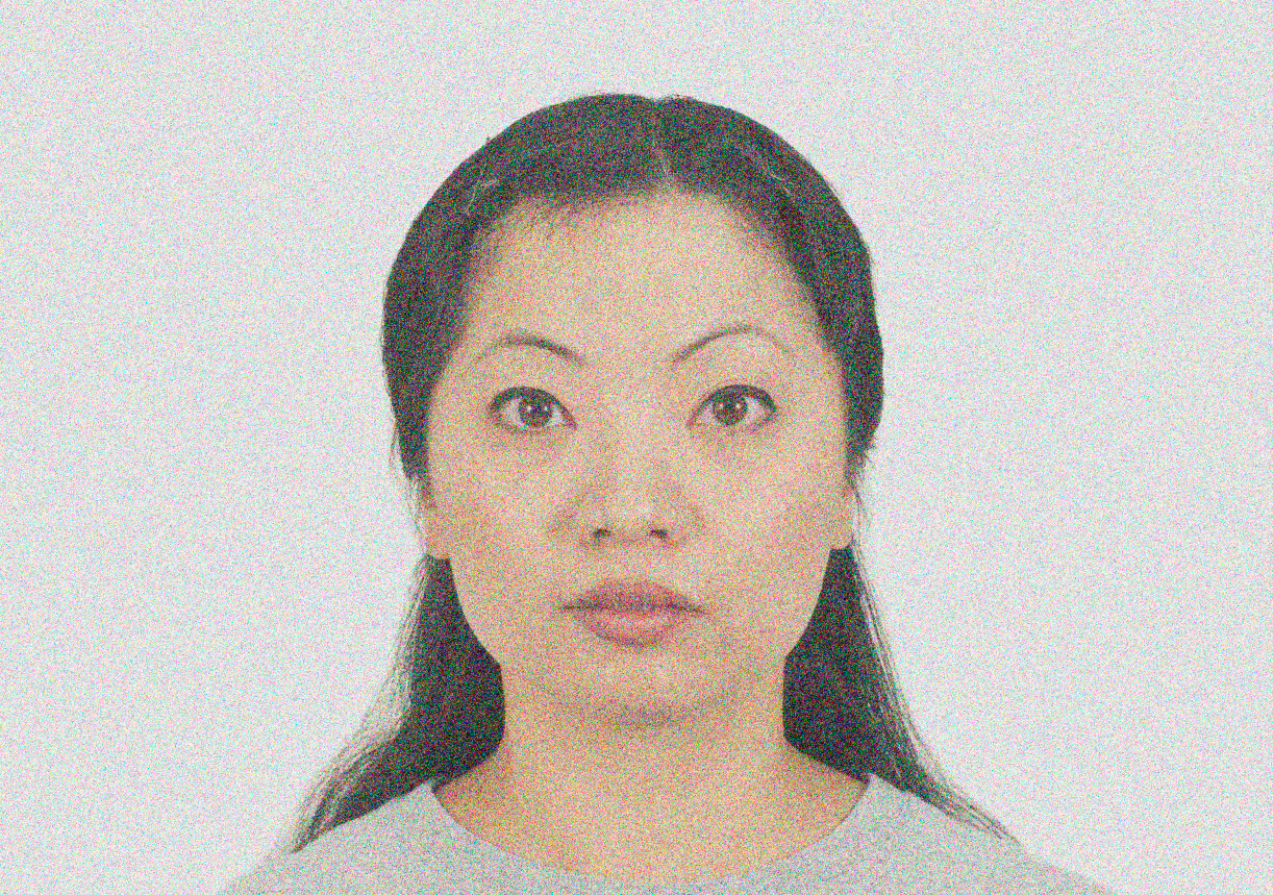}
		\caption{}
		\label{fig1:f}
	\end{subfigure}
	\caption{{\bf Images from \textsc{CelebSET} (a-c), \textsc{FairFace} (d-f) and \textsc{CFD} (g-i).}}
	\label{fig1}
	\vspace{-5 mm}
\end{figure}

In this section, we describe the datasets that we use for our experiments\footnote{Dataset and code are available at https://tinyurl.com/xt5wdmw} along with an outline of the perturbations that we have applied on the images to create adversarial inputs. 

\subsection{Datasets}
We use three annonated datasets for auditing the FRSs. These datasets are composed of face images of people from different demographics (based on ethnicity, gender, and age group): 
\textsc{CelebSET}~\cite{raji2020saving}, \textsc{Chicago Face Database}~\cite{ma2015chicago,ma2020chicago,lakshmi2021india} \blu{and \textsc{FairFace}~\cite{karkkainenfairface}.}

\vspace{1 mm}
\noindent
\textsc{CelebSET} -- 
This dataset is composed of 1600 images of 80 celebrities -- 20 of the most photographed celebrities from each of the following intersectional groups: White Male (WM), White Female (WF), Black Male (BM) and Black Female (BF)~\cite{raji2020saving}. Thus there are 20 celebrities from each category with 20 images each. The authors manually selected 10 smiling and 10 non-smiling images of each celebrity. The images are annotated with name, ethnicity, gender, age, smile and bounding box coordinates. The age of the person in the photo is estimated by subtracting the birth-date from the date of the photo. Each of the photos is present as a 128 $\times$ 128 image file with only the face area visible. The photos are \textbf{not} of very high quality 
and there is no consistency in the lighting or viewing angle. One of the images from this dataset is shown in Figure~\ref{fig1}(a).

\vspace{1 mm}
\noindent
\blu{\textsc{FairFace} -- 
This dataset is composed of 108,501 images taken from the YFCC-100M Flickr dataset belonging to the following ethnic groups -- White, Black, Latino, Indian, Southeast Asian, East Asian and Middle Eastern~\cite{karkkainenfairface}. The images are annotated with gender, age and ethnicity. We randomly sampled 7000 images from this dataset for our experiments. We chose 1000 images for each ethnicity equally divided for the two genders. Each picture is present as a 224 $\times$ 224 image file with only the face area visible. The photos are \textbf{not} of very high quality and there is no consistency in the lighting or viewing angle. One of the images from this dataset is shown in Figure~\ref{fig1}(d)}.

\vspace{1 mm}
\noindent
\textsc{Chicago Face Database(CFD)} -- 
This dataset is composed of 597 unique images of American citizens belonging to the following ethnic categories: White, Black, Asian and Latino~\cite{ma2015chicago}. The images are annotated with ethnicity, gender, age, emotion, facial landmarks, etc.\footnote{More details available at: \url{https://chicagofaces.org/default/}}. The distribution of males and females in the dataset for each of these ethnic groups is noted in Table~\ref{tab1}. These are high-resolution images with standardized lighting conditions and angle of view with a resolution of 2444 $\times$ 1718 on a white background. All images have a neutral face expression. A sample image from this dataset is shown in Figure~\ref{fig1}(g).
There are two more datasets we use that fall under the purview of this database. 
\begin{compactitem}
	\item CFD-MR~\cite{ma2020chicago} -- Composed of 88 unique images of American individuals who have parents belonging to different ethnic groups. 
	\item CFD-India~\cite{lakshmi2021india} -- Composed of 142 unique images of Indian individuals from New Delhi, India, annotated for the same features as above. 
\end{compactitem}

\begin{table}
    \small
	\begin{center}
		\begin{tabular}{l | c| c | c | c} 
			& White & Black & Asian & Latino \\
			\hline
			Males\rule{0pt}{2ex} & 93 & 93 & 52 & 52 \\
			\hline
			Females\rule{0pt}{2ex} & 90 & 104 & 57 & 56 \\
			\hline
		\end{tabular}
	\end{center}
	\caption{\textbf{Distribution of males and females across different ethnic groups in CFD dataset.}}
	\label{tab1}
	\vspace{-5 mm}
\end{table}

\subsection{Inputs for adversarial audits}
Recently, adversarial image manipulation techniques have been widely used to confuse face detection softwares successfully~\cite{vakhshiteh2020adversarial}. Even though there are multiple sophisticated deep learning tools available for generating such adversarial/perturbed images, in this work, we have made a conscious decision to step away from these sophisticated techniques. We have chosen simpler tools and techniques that are more accessible to people outside the academic community for perturbing the images. In particular, to perturb an image we use the open-source image manipulation program -- \textsc{GIMP}~\cite{gimp} to create adversarial images to test the FRSs. \textsc{GIMP} is freely available and has an intuitive GUI with a well-defined documentation and therefore easily accessible for users intending to apply digital filters or perform various image editing tasks. Our hypothesis is that even such simple techniques available to unassuming users can elicit biased responses from these softwares which can have dangerous ramifications. We highlight a few ways in which such perturbed images could potentially manifest in physical and digital world and can be nicely simulated by the \textsc{GIMP} software. 
\begin{itemize}
	\item\label{adv-item1} Social media applications like Instagram and Snapchat allow users to apply various simple types of digital filters on photos (during and after capturing). Such filters may distort or introduce various types of noises in the images to emulate the effect of old analog cameras. These edited photos are available to the back-end system as well as followers of the user and can be easily stored and used for training/testing of FRSs. To simulate this scenario we use the \textit{RGB} noise filter which adds a normally distributed noise to a layer or a selection, thereby, giving the effect of a grainy texture to an image. 
	\blu{This filter is applied to photos in all the datasets -- \textsc{CelebSET}, \textsc{FairFace} and \textsc{CFD} with examples in Figures~\ref{fig1} (c), (f) and (i) }
	
	\item CCTV cameras are being used increasingly for surveillance and policing of individuals \cite{singh_frt2021,marrow_frt2020}
	by various governments. These cameras are placed in public places and exposed to environmental elements like rain, dirt, etc. Sometimes, a single camera may be covering a large area and therefore needs to use zoom function which dilates or spreads the pixels of the images. These unintended changes can result in disastrous consequences as the facial recognition tasks are completely automated~\cite{lomas_frt2020}. In order to emulate this scenario we use the \textit{spread} noise filter which swaps each pixel in the active layer or selection with another randomly chosen pixel by a user specified amount, thereby, giving a slightly jittery output. 
	\blu{This filter is applied to photos in all the datasets -- \textsc{CelebSET}, \textsc{FairFace} and \textsc{CFD}, with one example in Figure~\ref{fig1}(b).}
	
	\item Physical photographs clicked on analog or digital cameras, or from old newspapers are restored to a digital format by scanning\footnote{\url{https://affidavit.eci.gov.in/candidate-affidavit}} and compressing. Such digital copies are neither high quality nor are they expected to be in any standardized format. Using FRSs to identify people from such images may result in biased outcomes
	. These biases may even increase due to the quality of the image. To simulate this situation, we use the \textit{newsprint} filter which halftones the image using a clustered-dot dither, thereby, giving the effect of a newspaper print. Given the suitability of the images based on their quality and on manual inspection of the output images, we determined that this filter should be applied to photos in \textsc{CFD} only, with one example shown in Figure~\ref{fig1}(h).

\end{itemize}

\if{0}All the above examples motivate a study to understand the change in bias in face analysis tasks for noisy/perturbed input images. To this end, we use an open-source image manipulation program -- \textsc{GIMP}\cite{gimp} to create adversarial images to test the FRSs. \textsc{GIMP} is freely available and has a intuitive GUI with a well-defined documentation and therefore easily accessible for users intending to apply digital filters or perform various image editing tasks.
Specifically, we use the following filters as part of our study.
\begin{enumerate}
	\item \textbf{RGB noise filter}: This filter adds a normally distributed noise to a layer or a selection, thereby, giving the effect of a grainy texture to an image. This filter is applied to photos in both \textsc{CelebSET} and \textsc{CFD} with examples in Figures~\ref{fig1} (c) and (f) and emulates the scenario described for social media apps above.
	\item \textbf{Spread noise filter}: This filter swaps each pixel in the active layer or selection with another randomly chosen pixel by a user specified amount, thereby, giving a slightly jittery output. This filter is applied to photos in \textsc{CelebSET} and \textsc{CFD}, with one example in Figure~\ref{fig1}(b). This filter attempts to emulate the scenario for CCTV cameras described above.
	\item \textbf{Newsprint filter:} This filter halftones the image using a clustered-dot dither, thereby, giving the effect of a newspaper print. Given the suitability of the images based on their quality and on manual inspection of the output images, we determined that this filter should be applied to photos in \textsc{CFD} only, with one example in Figure~\ref{fig1}(e) and emulates the effect of old newspaper prints of the photos as discussed above.
\end{enumerate}
\doubt{Will it be better if we combine the bulleted points with the list of filters and write them together?--AD}\fi

\section{Experiments and observations}
\label{sec:experiments}
In this section, we present the experimental setup followed by a discussion on the observations for both traditional and adversarial audits conducted on the aforementioned datasets.

\subsection{Tasks, datasets, and experimental setup}

In this study, we evaluate the FRSs 
for the following tasks.
\begin{compactitem}
	\item \textbf{Gender detection}: Each of the FRS report either \textit{Male} or \textit{Female} gender for an input image. 
	\item \textbf{Age prediction}: Given the reported age from Microsoft, Face++, and median of the age range of AWS, we use a 8 years acceptance margin, i.e., ground truth $\pm 4$. If the reported age falls in the margin of acceptance, we consider it to be a correct prediction. 
	\item \textbf{Smile detection}: For this task, AWS reports a `smile' or `no-smile' label and a confidence score for its prediction. Face++ and Microsoft report a confidence value in $\lbrack 0,100 \rbrack$ and $\lbrack 0,1 \rbrack$ respectively. Thus we consider the threshold as $50\%$ and $0.5$ respectively to consider the prediction as true for Face++ and Microsoft.
\end{compactitem}

\vspace{1 mm}
\noindent
\blu{\textbf{Datasets}: As mentioned earlier, we evaluate the following datasets -- \textsc{CelebSET}, \textsc{FairFace} and \textsc{CFD} (and its extensions -- \textsc{CFD-MR} and \textsc{CFD-India}). 
We do not evaluate the smile prediction task for FairFace (not annotated) and CFD datasets (all images have a neutral facial expression). 
A summary of the triplets of $<$dataset, filter, task$>$ that we evaluate is noted in Table~\ref{tab2:dataset-filter-tasks}.}
\begin{table}
	\small
	\begin{center}
		\begin{tabular}{ c | c | c} 
			\textbf{Dataset} & \textbf{Filters} & \textbf{Tasks} \\
			\hline
			\textbf{\textsc{CelebSET}}\rule{0pt}{2ex} & Spread $|$ RGB & Gender $|$ Age $|$ Smile\\
			\hline
			\textbf{\textsc{FairFace}}\rule{0pt}{2ex} & Spread $|$ RGB & Gender $|$ Age \\
			\hline
			\textbf{\textsc{CFD}}\rule{0pt}{2ex} & Newsprint $|$ Spread $|$ RGB & Gender $|$ Age \\
			\hline
		\end{tabular}
	\end{center}
	\vspace{- 2 mm}
	\caption{\blu{\textbf{Filters and tasks for each dataset.}}}\label{tab2:dataset-filter-tasks}
	\vspace{- 2mm}
\end{table}

\vspace{1 mm}
\noindent
\textbf{Experimental setup}: We evaluate all datasets in batch mode through API calls to the FRS servers.
The parameter settings for each filter on \textsc{Gimp} are as follows.
\begin{compactitem}
	\item \textbf{Spread}: Horizontal/vertical spread amount -- 50 units for \textsc{CFD} and 8 for \textsc{CelebSET} \blu{and \textsc{FairFace}}.
	\item \textbf{Newsprint}: Screen cell width -- 20 pixels, \% of black pulled out -- $12\%$, screen angle for all colors -- 75$^{\circ}$.
	\item \textbf{RGB}: Noise in RGB channels -- 0.8 for \textsc{CFD} and 0.5 for \textsc{CelebSET} \blu{and \textsc{FairFace}}.
\end{compactitem}

With higher noise values, images in \textsc{CelebSET} \blu{and \textsc{FairFace}} were not recognisable to human eyes either. Given the smaller resolution, we have intentionally included less noise to these images. As mentioned in the dataset section, \textsc{CelebSET} \blu{and \textsc{FairFace}} images were heavily distorted due to newsprint filter too. Hence, no further experiments were done for that setting.

\subsection{Observations for \textsc{CelebSET}}
We first report the observations of the traditional audit experiments followed by the results of adversarial audits.
 
\begin{table}
	\small
	\begin{center}
		\begin{tabular}{c | c | c | c} 
			\hline
			FRS\rule{0pt}{2ex} & Gender & Age & Smile \\ 
			\hline
			AWS\rule{0pt}{2ex} & $-0.25\%$& $-17.46\%$& $-5.54\%$ \\ 
			Microsoft\rule{0pt}{2ex} & $-0.63\%$ & $-21.22\%$ & $+12.94\%$ \\
			\hline
		\end{tabular}
	\end{center}
	\vspace{- 2 mm}
	\caption{\label{tab:celebset-old-new}\textbf{Change in accuracy for \textsc{CelebSET} relative to Saving Face (\citet{raji2020saving}}).}
	\vspace{-6 mm}
\end{table}

\subsubsection{Change in prediction accuracy over time}
Table~\ref{tab:celebset-old-new} shows the change in accuracy for the gender, age, and smile prediction tasks as compared against the findings of ~\citet{raji2020saving}. We report the difference between the accuracy obtained in our investigation and that reported by Raji et al. Hence a $-ve$ sign denotes that for the corresponding FRS, for the corresponding task the prediction accuracy has decreased compared to what was reported earlier.

\noindent
\textbf{Observations}: Both AWS and Microsoft report a marginal (insignificant) drop in accuracy for gender detection by $0.25\%$ and $0.63\%$ respectively. However, the drop in accuracy for age prediction is alarming with a fall of more than 15\% for both the FRS. 
Interestingly, while AWS performance has also deteriorated in smile detection task, Microsoft has actually improved by a significant margin ($13\%$).

\noindent
\textbf{Takeaways}: While the FRS have somewhat stable performance in the gender detection task, we observe significant difference in performance in the other two tasks. This could be possibly an outcome of the fact that fixing of results in one dimension might adversely affect the performance in another dimension. However one needs to have access to the black box algorithms to corroborate this postulation.    
\subsubsection{Effect of perturbations on the performance}

\begin{figure*}[t]
	\centering
	\begin{subfigure}{0.66\columnwidth}
		\includegraphics[width= \textwidth, height=3.5cm, keepaspectratio]{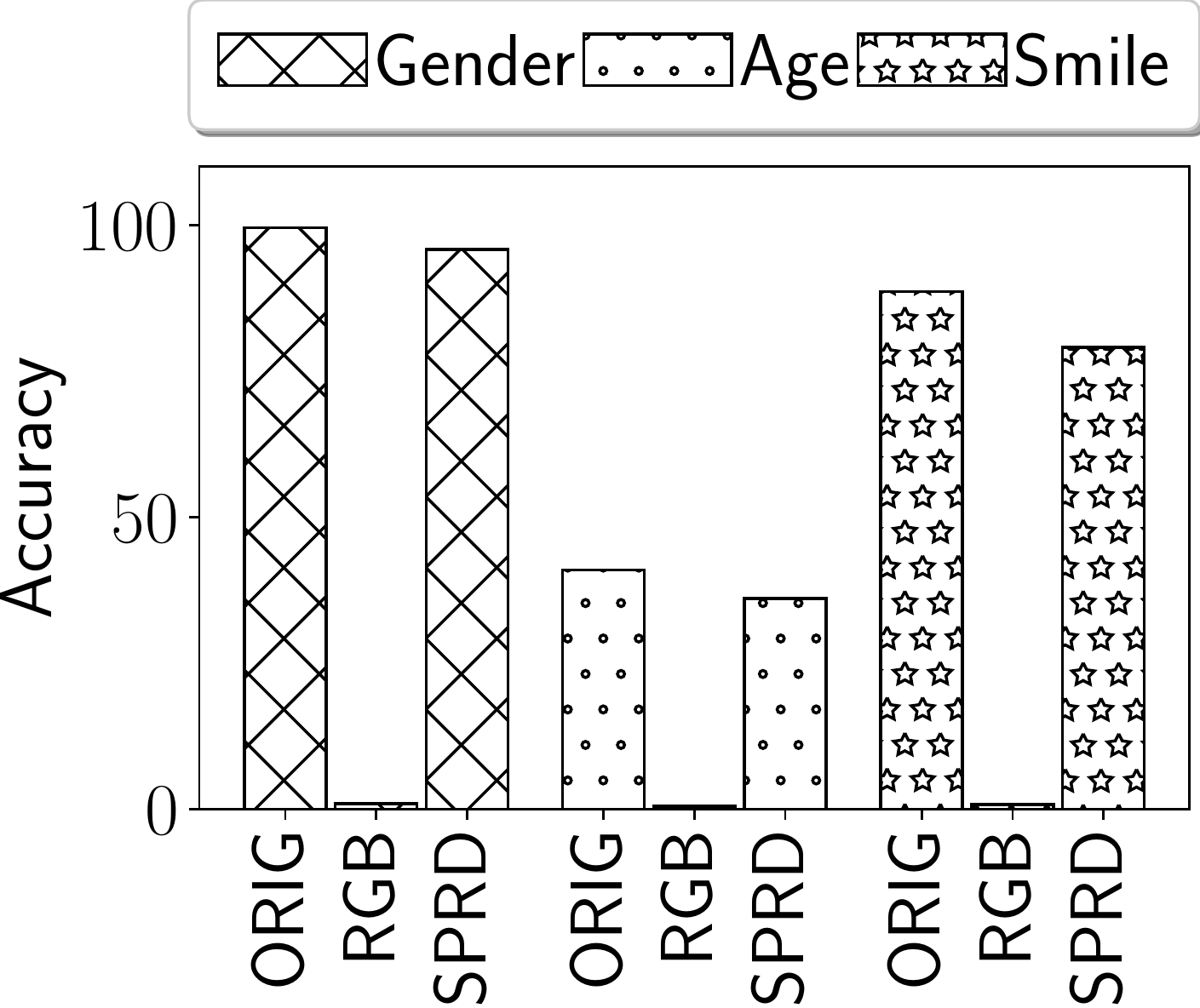}
		\caption{Accuracy on AWS}
		\label{Fig: AWS-celebset-accuracy}
	\end{subfigure}%
	~\begin{subfigure}{0.66\columnwidth}
		\includegraphics[width= \textwidth, height=3.5cm, keepaspectratio]{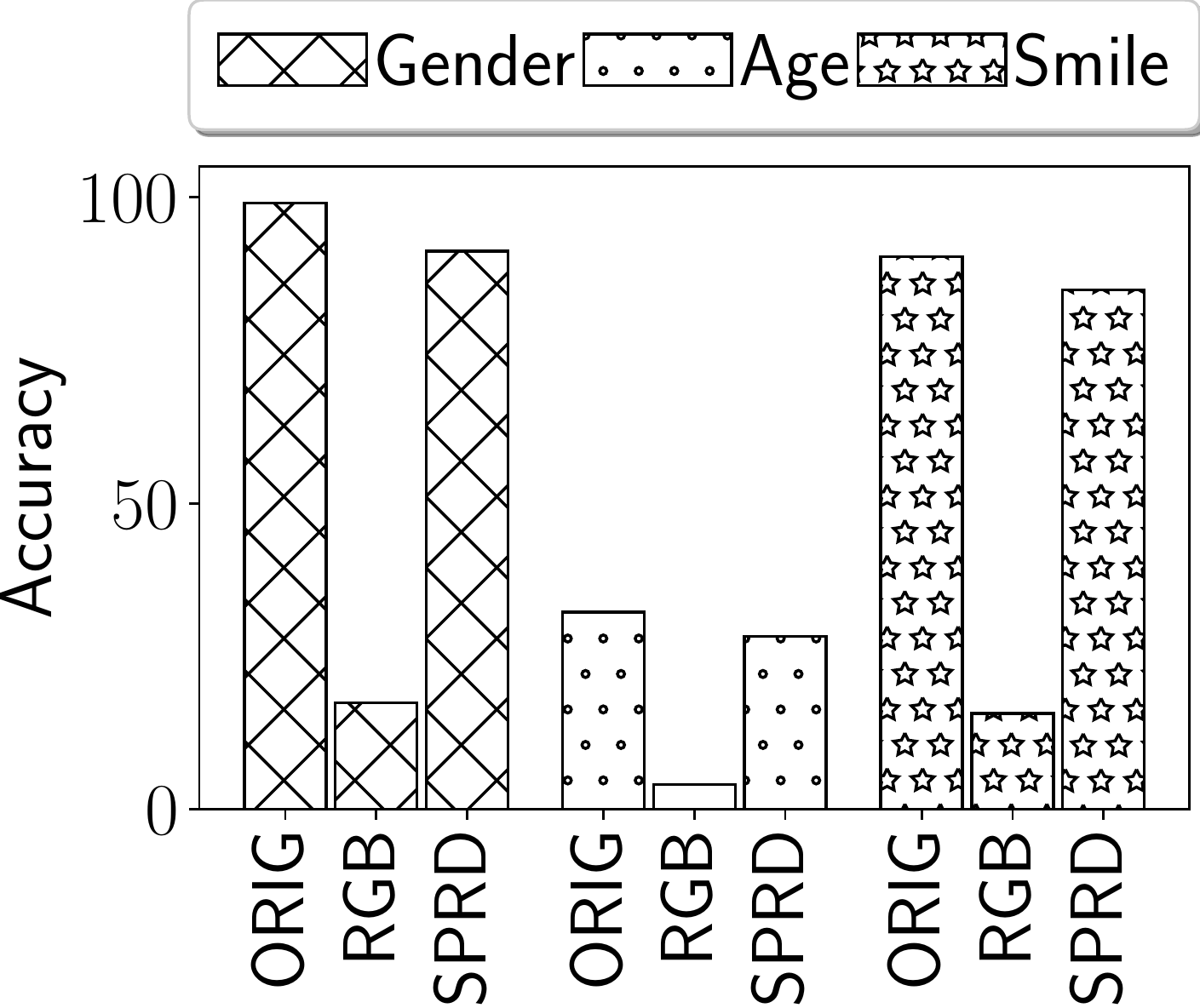}
		\caption{Accuracy on Face++}
		\label{Fig: FPP-celebset-accuracy}
	\end{subfigure}
	~\begin{subfigure}{0.66\columnwidth}
		\includegraphics[width= \textwidth, height=3.5cm, keepaspectratio]{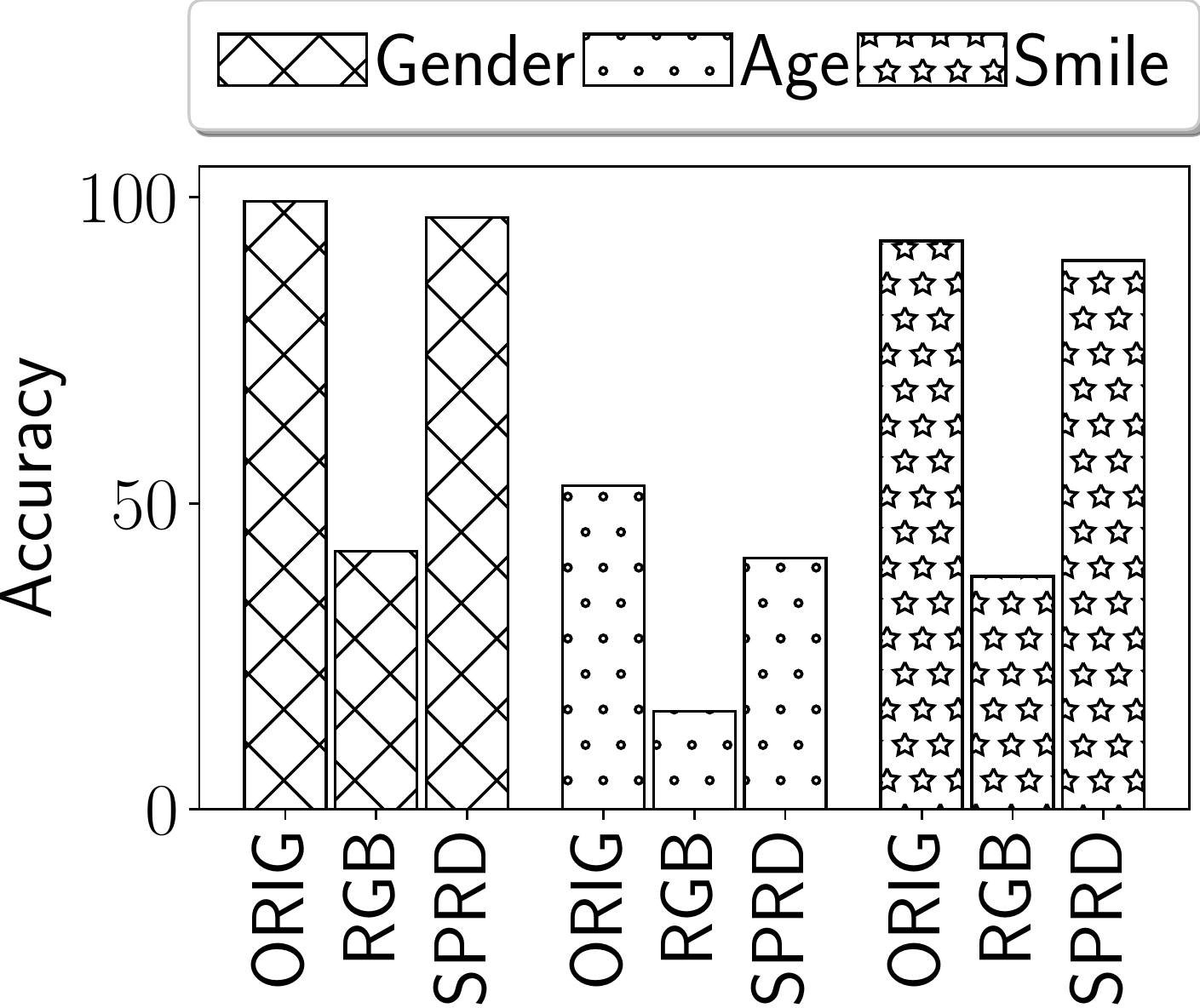}
		\caption{Accuracy on Microsoft}
		\label{Fig: MSFT-celebset-accuracy}
	\end{subfigure}
	
	\caption{{\bf Prediction accuracy on \textsc{CelebSET} for original, RGB-filter and spread-filter inputs. Under adversarial settings, all the FRSs suffer from drop in accuracy (AWS has $\bf < 2\%$ accuracy), with Microsoft being marginally more robust than AWS and Face++.}}
	\label{fig:celebset-accuracy}
	\vspace{-4 mm}
\end{figure*}

\begin{table*}[!ht]
	\small
	\begin{center}
		\begin{tabular}{c | c | c | c} 
			\hline
			\textbf{Image}\rule{0pt}{2ex} & \textbf{Gender} & \textbf{Age} & \textbf{Smile}\\
			\hline
			AWS/original\rule{0pt}{2ex} & \textbf{(WF - WM)} $1\%$ & \textbf{(WF - BF)} $19.25\%$ & \textbf{(BF - WM)} $6.5\%$ \\ 
			AWS/RGB\rule{0pt}{2ex} & \textbf{(WM - BM/BF)} $1.75\%$ & \textbf{(WM - BM)} $1.25\%$ & \textbf{(BF - BM)} $1.25\%$ \\
			AWS/spread\rule{0pt}{2ex} & \textbf{(WF - WM)} $\textbf{5.5\%}$ & \textbf{(WF - BF)} $\textbf{19.5\%}$ & \textbf{(BF - WM)} $\textbf{12.5\%}$ \\
			\cellcolor{red!20}Max. absolute increase in disparity\rule{0pt}{2ex} & \cellcolor{red!20} \textbf{4.5\% (spread)} & \cellcolor{red!20} \textbf{0.5\% (spread)} & \cellcolor{red!20} \textbf{6\% (spread)}\\
			\hline
			Face++/original\rule{0pt}{2ex} & \textbf{(WM - BF)} $2.5\%$ & \textbf{(WF - WM)} $\textbf{13.5\%}$ & \textbf{(BF - BM)} $7\%$ \\ 
			Face++/RGB\rule{0pt}{2ex} & \textbf{(WF - BM)} $\textbf{22.75\%}$ & \textbf{(BF - BM)} $4.25\%$ & \textbf{(WF - BM)} $\textbf{20.75\%}$ \\
			Face++/spread\rule{0pt}{2ex} & \textbf{(WF - BM)} $17.5\%$ & \textbf{(BM - WM)} $2.5\%$ & \textbf{(WF - BM)} $12.5\%$ \\
			\cellcolor{red!20}Max. absolute increase in disparity\rule{0pt}{2ex} & \cellcolor{red!20} \textbf{20.25\% (RGB)} & \cellcolor{red!20} $\phi$ & \cellcolor{red!20} \textbf{13.75\% (RGB)}\\
			\hline
			Microsoft/original\rule{0pt}{2ex} & \textbf{(WM/WF - BM/BF)} $1\%$ & \textbf{(BM - BF)} $16\%$ & \textbf{(WF - BM)} $4.5\%$ \\ 
			Microsoft/RGB\rule{0pt}{2ex} & \textbf{(WF - BM/BF)} $\textbf{36.25\%}$ & \textbf{(WF - BM)} $15.75\%$ & \textbf{(WF - BM)} $\textbf{33.75\%}$ \\
			Microsoft/spread\rule{0pt}{2ex} & \textbf{(WF - BM/BF)} $6.5\%$ & \textbf{(WM - BF)} $\textbf{30\%}$ & \textbf{(WF - BM)} $9.5\%$ \\
			\cellcolor{red!20}Max. absolute increase in disparity\rule{0pt}{2ex} & \cellcolor{red!20} \textbf{35.25\% (RGB)} & \cellcolor{red!20} \textbf{14\% (spread)} & \cellcolor{red!20} \textbf{29.25\% (RGB)}\\
			\hline
		\end{tabular}
		\caption{\label{tab:celebset-intersection}\textbf{Difference in highest accuracy and lowest accuracy amongst the intersectional groups for \textsc{CelebSET}. Most results show significant disparity in prediction accuracy for people of `Black' ethnicity (BF/BM) as compared to that of `White'  ethnicity. Microsft reports the largest disparity for all tasks, irrespective of the input.}}
	\end{center}
	\vspace{-5 mm}
\end{table*}

The first striking observation that we make is on the number of images identified by the FRSs.
\begin{compactitem}
	\item \textbf{Original} -- AWS: 1600, Face++: 1600, Microsoft: 1590.
	\item \textbf{Spread Filter} -- AWS: 1549, Face++: 1534, Microsoft: 1552.
	\item \textbf{RGB Filter} -- AWS: 20, Face++: 318, Microsoft: 683.
\end{compactitem}

The above list shows that adding adversarial filters to low quality images successfully fools the facial recognition systems. It is interesting to note that the RGB filter has an extremely high mis-classification rate for all FRSs with faces identified on less than $2\%$ images for AWS. Figure \ref{fig:celebset-accuracy} presents the prediction accuracy for the three tasks of gender, age and smile detection. Each of the plots Figures \ref{Fig: AWS-celebset-accuracy}, \ref{Fig: FPP-celebset-accuracy} and \ref{Fig: MSFT-celebset-accuracy} correspond to an FRS API. 


\noindent
\textbf{Gender identification}: We see that all the systems report high gender identification accuracy ($\geq 99\%$) for the original dataset. However, with the addition of noise (i.e., in adversarial setting), this accuracy is observed to be significantly affected. 
AWS reports an accuracy of only $1.5\%$ for the RGB filter (Figure \ref{Fig: AWS-celebset-accuracy}). The drop in accuracy for spread filter is not as significant with the lowest reported accuracy being $92\%$ for Face++ (Figure \ref{Fig: FPP-celebset-accuracy}).

\noindent
\textbf{Age prediction}: For the task of age prediction, all APIs perform badly even for the original input with only Microsoft having an accuracy of $ > 50\%$ (Figure \ref{Fig: MSFT-celebset-accuracy}). The adversarial images further reduce the performance with RGB filter causing a drop of $ > 50\%$. Here again, Face++ has the lowest accuracy for the spread filter --  $\approx 28\%$ (Figure \ref{Fig: FPP-celebset-accuracy}).

\noindent
\textbf{Smile detection}: All APIs are able to identify smiles with an accuracy $\geq 88\%$ for the original dataset, but exhibit a similar behaviour for the accuracy on RGB and spread filters, with Microsoft being the best overall and AWS being the worst. 

\noindent
The summary of the findings from Figure \ref{fig:celebset-accuracy} are as follows.
\begin{compactitem}
	\item FRSs perform well on the original input for gender and smile detection tasks, but for age detection their accuracy drops significantly.
	\item The perturbation caused due to different noises seem to affect the performance of the FRSs. Especially, RGB noise has significant deterioration effect on the performance across all the systems.
	\item Microsoft is the most robust FRS across all tasks and inputs among the three systems. However, its performance is also badly affected by the RGB noise.
\end{compactitem}

%
%
%

\subsubsection{Disparity in accuracy for intersectional groups}
We now evaluate the disparity in accuracy by taking the difference of the highest accuracy and the lowest accuracy among the different intersectional groups -- WM, WF, BM, BF of \textsc{CelebSET} dataset. The results are shown in Table~\ref{tab:celebset-intersection}. Each cell in the table denotes the highest disparity for a given task for the corresponding dataset. The mentions within parenthesis denote the intersectional groups between which the corresponding disparity was observed. Rows marked in red denote the maximum absolute increase in disparity from original to any of the two perturbed images.

\noindent
\textbf{Gender detection}: For the original images, we do \textbf{not} observe any significant disparities between performance on intersectional groups ($\le 3\%$ across all systems). The disparity increases for the two adversarial filters for all facial recognition systems. The maximum disparity of $36.25\%$ is observed for Microsoft for RGB noise. This disparity is mostly against people of `Black' ethnicity as compared to people of `White' ethnicity. In fact, this disparity is highest for Microsoft despite it being the most accurate (overall) amongst the three systems (see Figure \ref{fig:celebset-accuracy}). It is interesting to note that even though spread noise did not have any significant effect on the overall accuracy, the disparity among intersectional groups saw significance rise in all FRSs (as high as 17.5\% for Face++).

\noindent
\textbf{Age prediction}: For the original images, we observe significant disparities between performance on intersectional groups ($\ge 10\%$ across all systems). Disparity in accuracy in case of age prediction shows some intriguing trends. For RGB noise, we observe that the disparity has decreased in all FRSs. However, that is mostly due to the significant drop in overall accuracy in all systems. However, for spread noise, we observe that the disparity in Microsoft system has increased to 30\% (an increase of 14\%). Interestingly, for the rest of the two systems, we do not observe significant rise in disparity. In fact, the disparity reduces in Face++ by almost 11\%.


\noindent
\textbf{Smile detection}: For the original images, we observe marginal disparities between performance on intersectional groups. 
However, the disparity in accuracy increases upon addition of noise in the images (only aberration is -- AWS/RGB). We observe that both Face++ and Microsoft report the highest disparity for RGB filter and AWS does so for the spread filter. This is to the extent that the disparity in Microsoft for smile detection actually becomes 7.5 times of what it is for original images.

\noindent
We observe similar trends for biases between gender and ethnic groups taken together. Results have been omitted for brevity.

\noindent
\textbf{Summary}
\begin{compactitem}
	\item All FRSs exhibit significant increase in disparity of accuracy for both gender and smile detection upon introduction of noise in the original images.
	\item The observed disparity in accuracy is against individuals of `Black' ethnicity (BM or BF).
	\item Even though we found Microsoft system to be the most robust among the three, the disparities between the intesectional groups are also the maximum. Thus the high accuracy is not dispersed among all intersectional groups uniformly; rather error rate is significantly higher for images of dark skinned people.
\end{compactitem}
%
%
%

\blu{\subsection{Observations for \textsc{FairFace}}}
\blu{We perform similar experiments as \textsc{CelebSET}, on \textsc{FairFace}. 
We observe that all FRSs miss out on identifying a certain number of faces in the original set of images. This is further exacerbated with the perturbed images, with Microsoft identifying the least number of faces.}
\purpl{
The number of images identified per FRS is
\begin{compactitem}
	\item \textbf{Original} -- AWS: 6903, Face++: 6918, Microsoft: 5717.
	\item \textbf{RGB Filter} -- AWS: 6229, Face++: 2461, Microsoft: 2163.
	\item \textbf{Spread Filter} -- AWS: 6855, Face++: 6764, Microsoft: 5647.
\end{compactitem}
} 

\begin{figure*}[t]
	\centering
	\begin{subfigure}{0.66\columnwidth}
		\includegraphics[width= \textwidth, height=3.5cm, keepaspectratio]{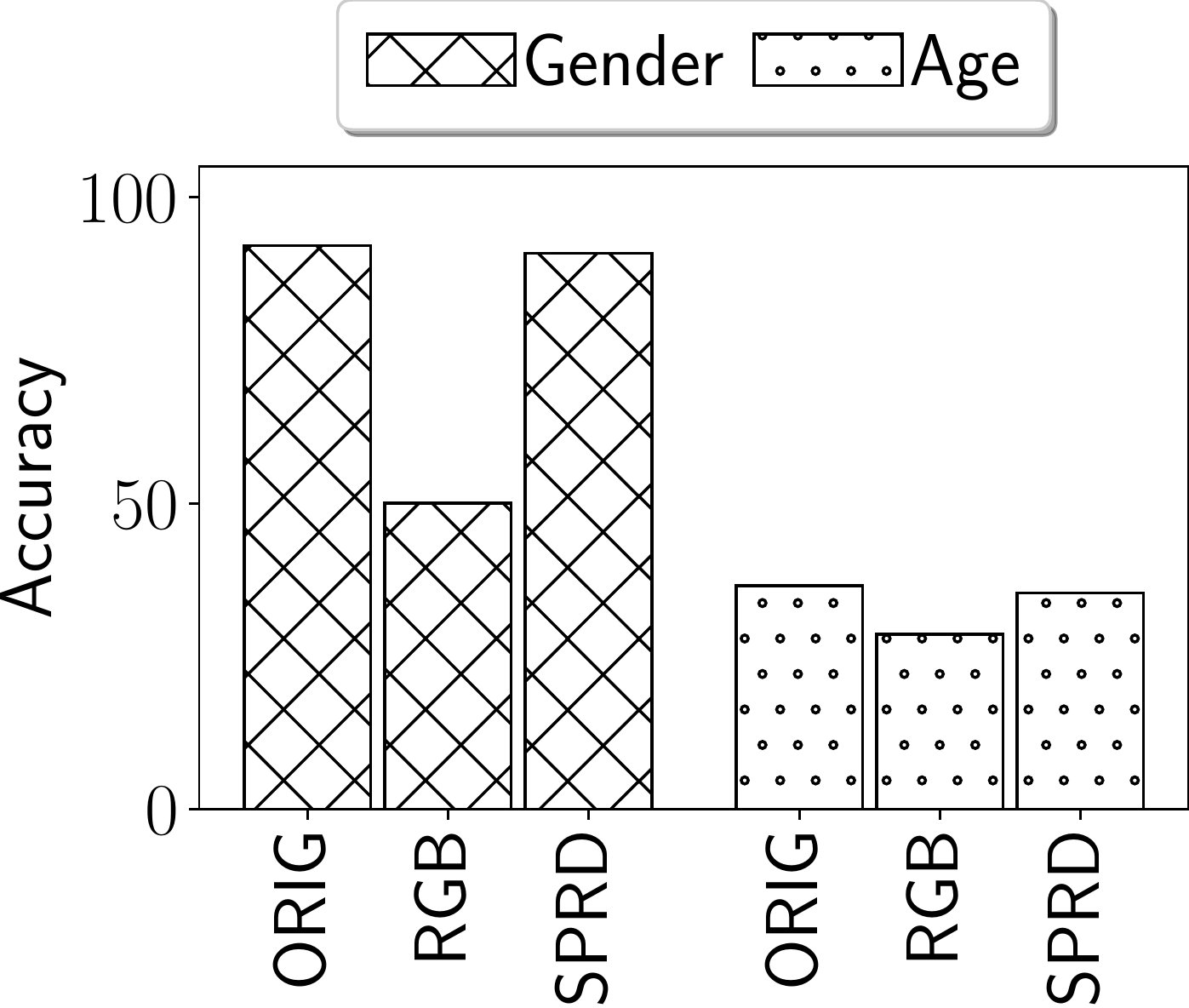}
		\caption{Accuracy on AWS}
		\label{Fig: AWS-fface-accuracy}
	\end{subfigure}%
	~\begin{subfigure}{0.66\columnwidth}
		\includegraphics[width= \textwidth, height=3.5cm, keepaspectratio]{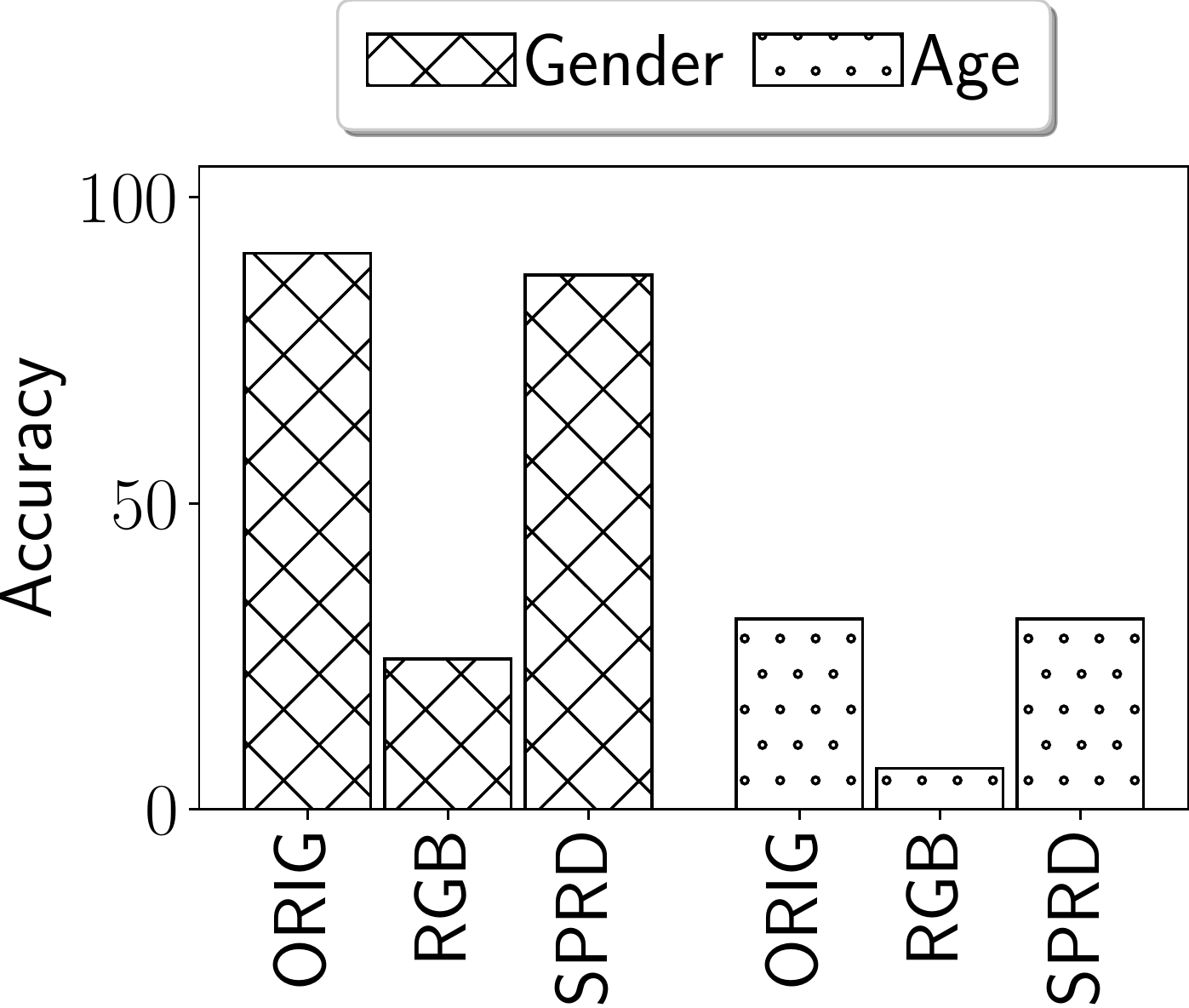}
		\caption{Accuracy on Face++}
		\label{Fig: FPP-fface-accuracy}
	\end{subfigure}
	~\begin{subfigure}{0.66\columnwidth}
		\includegraphics[width= \textwidth, height=3.5cm, keepaspectratio]{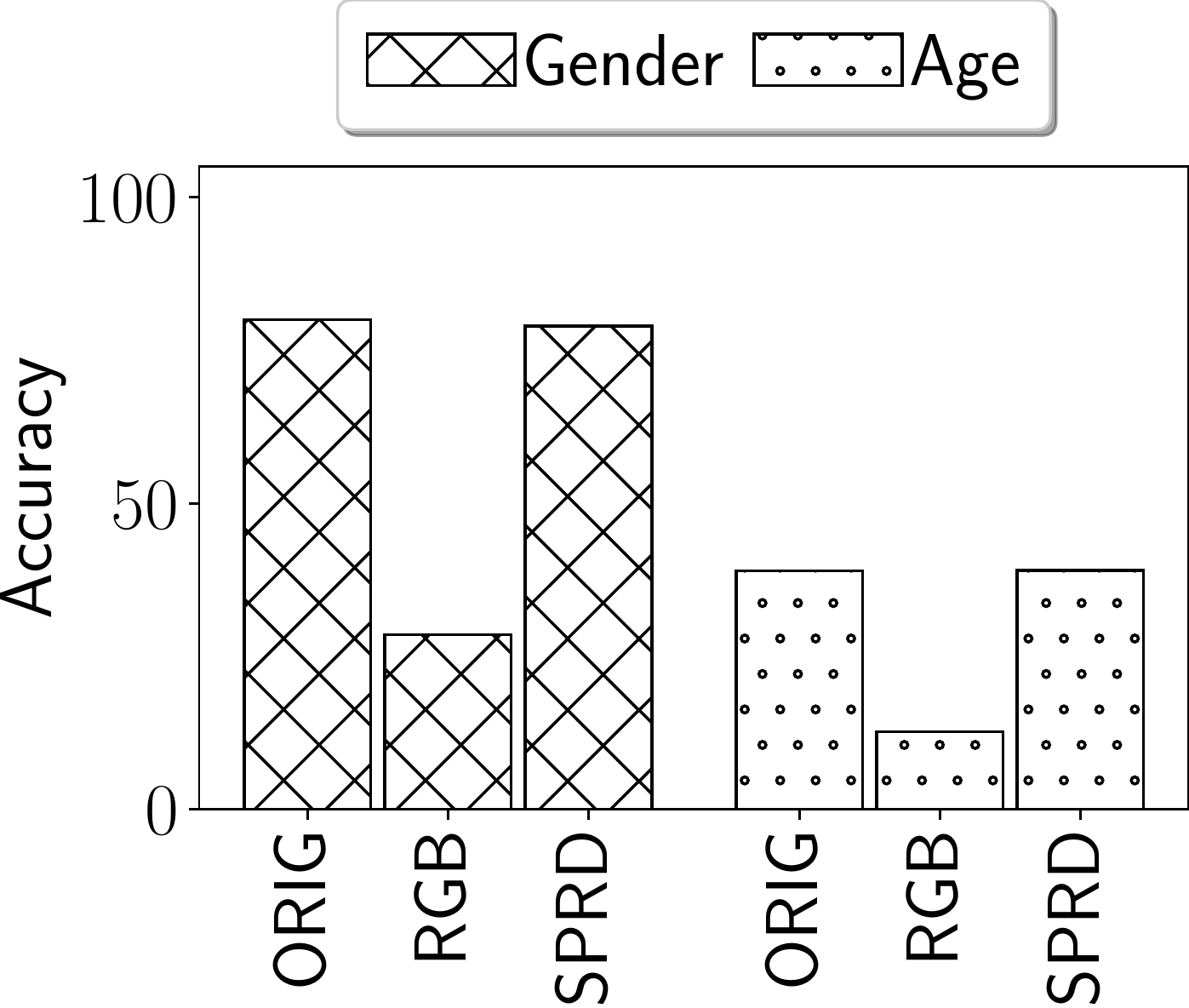}
		\caption{Accuracy on Microsoft}
		\label{Fig: MSFT-fface-accuracy}
	\end{subfigure}
	
	\caption{{\blu{\bf Prediction accuracy on \textsc{FairFace} for original, RGB-filter and spread-filter inputs. Under RGB and Spread adversarial settings, all of the FRSs suffer from drop in accuracy (Face++ accuracy drops by more than $\bf 65\%$), with AWS being significantly more robust than Microsoft and Face++.}}}
	\label{fig:fface-accuracy}
	\vspace{-4 mm}
\end{figure*}

\subsubsection{\blu{Effect of perturbations in the performance}} \blu{The results for the accuracy on gender and age detection tasks are presented in Figure \ref{fig:fface-accuracy} for original and adversarial images. 
} 

\noindent
\blu{\textbf{Gender detection}:}
\blu{ All the FRS APIs report good performance for gender detection. Microsoft API is found to be the worst among the three, identifying the gender correctly for $80\%$ of the images. On adding the perturbations, we see significant drop in accuracy for the RGB noise. We record a drop of more than $ 40\%$ for all FRSs. 
\purpl{Face++ and Microsoft are found to be the worst affected APIs with a drop of $ 66.26\%$ (Figure \ref{Fig: FPP-fface-accuracy}) and $51.45\%$ (Figure \ref{Fig: MSFT-fface-accuracy}) in accuracy respectively}. Similar to the \textsc{CelebSET} dataset, we notice that the FRSs are robust toward spread perturbation with no significant drop in accuracy.}

\noindent
\blu{\textbf{Age prediction}:}
\blu{We observe that the age prediction accuracy is poor ($\le 50\%$) for all APIs. The RGB noise perturbations show similar drop in accuracy for this task (as in gender detection). Face++ is observed to be the worst performing FRS with the \purpl{lowest age prediction accuracy of $6.7\%$ (Figure \ref{Fig: FPP-fface-accuracy}), followed closely by Microsoft at $ 12.67\%$ (Figure \ref{Fig: MSFT-fface-accuracy})}. For the spread noise, the drop is not significant; rather the accuracy for Microsoft API increases marginally.}

\vspace{1 mm}
\noindent
\blu{The key takeaways from Figure~\ref{fig:fface-accuracy} are as follows.}\blu{
\begin{compactitem}
	\item All APIs perform well for the task of gender detection for all inputs except RGB. However, the performance for age prediction is bad for all inputs on all APIs with Face++ and Microsoft being the worst.
	\item Microsoft is the worst performing FRS for gender detection on the original images while Face++ is the worst for age prediction. 
	\item On the RGB perturbed images, Face++ is the worst performing FRS for both tasks while AWS is the most robust.
    \item AWS is the most robust FRS among the three.
\end{compactitem}}

\subsubsection{\blu{Disparity in accuracy for intersectional groups}}

\blu{We evaluate the disparity in accuracy by taking the difference of the highest accuracy and the lowest accuracy among the different intersectional groups (two for each of the seven ethnic groups) -- WM, WF, BM, BF, IM, IF, LM, LF, SM, SF, EM, EF, MM and MF of \textsc{FairFace} dataset\footnote{\blu{Read the abbreviations WM: `White' males, WF: `White' females and so on for the different ethnic groups.}}. The results are presented in Table~\ref{tab:fface-intersection}. In the table, each cell denotes the highest disparity in accuracy for a given task. The phrases within the parenthesis correspond to the intersectional groups which are the farthest apart in accuracy. Finally, the rows marked in red show the absolute increase in disparity from the original images to any of the noises.}

\noindent
\blu{\textbf{Gender detection}:}
\blu{For the original input images, the disparity in accuracy is generally against the `Black' ethnicity with \purpl{Microsoft being the highest at 
$20.2\%$}. In the adversarial setting, for both Face++ and Microsoft, \purpl{the disparity is seen to be against `Black' and `Indian' ethnic groups}; whereas for AWS, it is against `East Asian' and `Southeast Asian' males. We notice that for the RGB noise, the disparity increases by more than 1.5 times for all FRSs with \purpl{AWS reporting a disparity of $73.6\%$ against `East Asian' males}. The maximum absolute increase in disparity is $63.60\%$ and reported for the RGB perturbation. \purpl{Although the disparities are distributed across ethnic groups, all are against people of a specific skin tone (`Asians', `Indians' or `Blacks').}}

\noindent
\blu{\textbf{Age prediction}:}
\blu{ For the task of age prediction, all the FRSs have a disparity of more than $9\%$ for original images. Here, the disparity is mostly against males, with \purpl{Microsoft reporting the highest disparity of $18.80\%$ against `Black' males. The disparity increases significantly for the RGB filter for AWS ($25\%$ against `Middle Eastern' males)} while it reduces for the other two FRSs. We notice that even with the spread filter, the disparity does not change by much and in fact reduces for Microsoft. \purpl{The most important observation here is that the disparity is against males for most of the cases (either `Middle Eastern' or `Black'), both from ethnic groups of color.}}

\noindent
\blu{\textbf{Summary}}
\blu{
\begin{compactitem}
	\item All FRSs exhibit an increase in disparity of accuracy for gender detection, mostly directed toward individuals of `Black' or `Asian' ethnicity.
	\item \purpl{AWS reports the highest disparity for all experiments with a value of $73.60\%$ between `Latino' females and `East Asian' males.}
	\item Microsoft reports the least absolute increase in disparity yet it is the most disparate against `Black' males; all its lowest accuracy values are reportedly for this group.
\end{compactitem}
}

\begin{table*}[!ht]
	\small
	\begin{center}
		\begin{tabular}{c | c | c } 
			\hline
			\textbf{Image}\rule{0pt}{2ex} & \textbf{Gender} & \textbf{Age} \\
			\hline
			AWS/original\rule{0pt}{2ex} & \textbf{(MF - SM)} $10\%$& \textbf{(EF - IM)} $14\%$ \\ 
			AWS/RGB\rule{0pt}{2ex} & \textbf{(EF - EM)} $\textbf{73.60\%}$ & \textbf{(EF - MM)} $\textbf{25\%}$ \\
			AWS/spread\rule{0pt}{2ex} & \textbf{(MF - EM)} $14\%$ & \textbf{(EF - IM)} $11.80\%$ \\
			\cellcolor{red!20}Max. absolute increase in disparity\rule{0pt}{2ex} & \cellcolor{red!20} \textbf{63.60\% (RGB)} & \cellcolor{red!20} \textbf{11\% (RGB)}\\		
			\hline
			Face++/original\rule{0pt}{2ex} & \textbf{(LM - BF)} $15.80\%$ & \textbf{(IM - MF)} $9.40\%$ \\ 
			Face++/RGB\rule{0pt}{2ex} & \textbf{(LF - BM)} $\textbf{30.60\%}$ & \textbf{(LF - BM)} $6.40\%$ \\
			Face++/spread\rule{0pt}{2ex} & \textbf{(MM - IF)} $14\%$ & \textbf{(SM - WF)} $\textbf{13.40\%}$ \\
			\cellcolor{red!20}Max. absolute increase in disparity\rule{0pt}{2ex} & \cellcolor{red!20} \textbf{14.80\% (RGB)} & \cellcolor{red!20} \textbf{4\% (spread)}\\
			\hline
			Microsoft/original\rule{0pt}{2ex} & \textbf{(LF - BM)} $20.20\%$& \textbf{(EF - BM)} $\textbf{18.80\%}$ \\ 
			Microsoft/RGB\rule{0pt}{2ex} & \textbf{(SF - BM)} $\textbf{30.60\%}$ & \textbf{(SF - BM)} $16.60\%$ \\
			Microsoft/spread\rule{0pt}{2ex} & \textbf{(LF - BM)} $22.40\%$ & \textbf{(EM - BM)} $\textbf{18.80\%}$ \\
			\cellcolor{red!20}Max. absolute increase in disparity\rule{0pt}{2ex} & \cellcolor{red!20} \textbf{10.40\% (RGB)} & \cellcolor{red!20} $\phi$\\
			\hline
		\end{tabular}
	\end{center}
	\caption{\label{tab:fface-intersection}\blu{\textbf{Difference in highest accuracy and lowest accuracy among the intersectional groups for \textsc{FairFace}. Most results show significant disparity in prediction accuracy for people of `Black' ethnicity (BF/BM) as compared to that of the other ethnic groups. AWS reports the largest disparity for all tasks on the RGB adversarial input.}}}
	\vspace{- 5 mm}
\end{table*}

\subsection{Observations for \textsc{CFD}}
We also perform a similar audit as \textsc{CelebSET}, on \textsc{CFD}. This dataset has higher quality images with standardized lighting and face angles and has individuals belonging to four different ethnic groups -- `White', `Black', `Latino', `Asian'. AWS and Microsoft APIs are able to identify faces in all 597 images but Face++ identifies only 423 images with faces for the newsprint filter.

%

\subsubsection{Effect of perturbations in the performance}
\begin{figure*}[t]
	\centering
	\begin{subfigure}{0.66\columnwidth}
		\includegraphics[width= \textwidth, height=3.5cm, keepaspectratio]{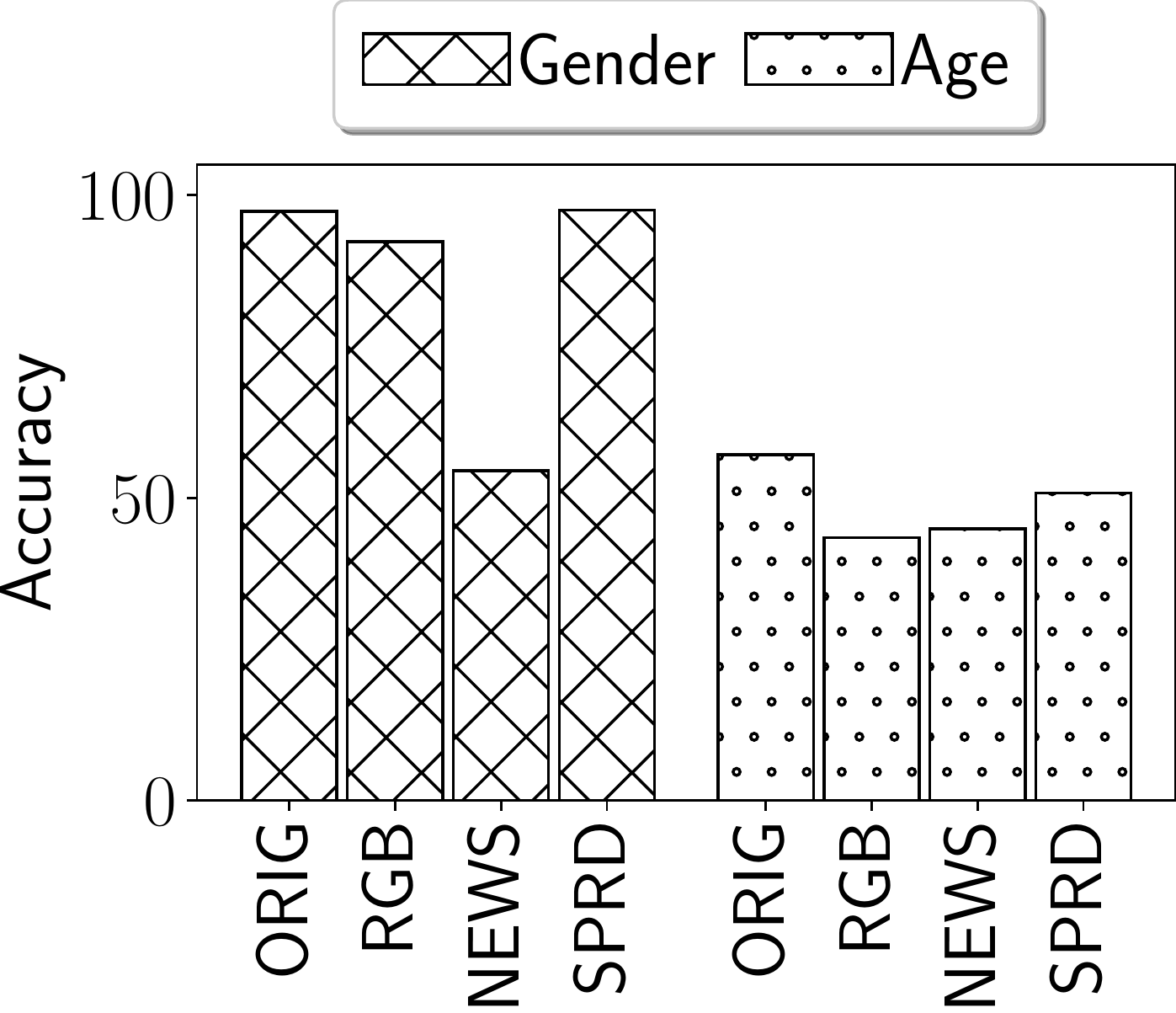}
		\caption{Accuracy on AWS}
		\label{Fig: AWS-cfd-accuracy}
	\end{subfigure}%
	~\begin{subfigure}{0.66\columnwidth}
		\includegraphics[width= \textwidth, height=3.5cm, keepaspectratio]{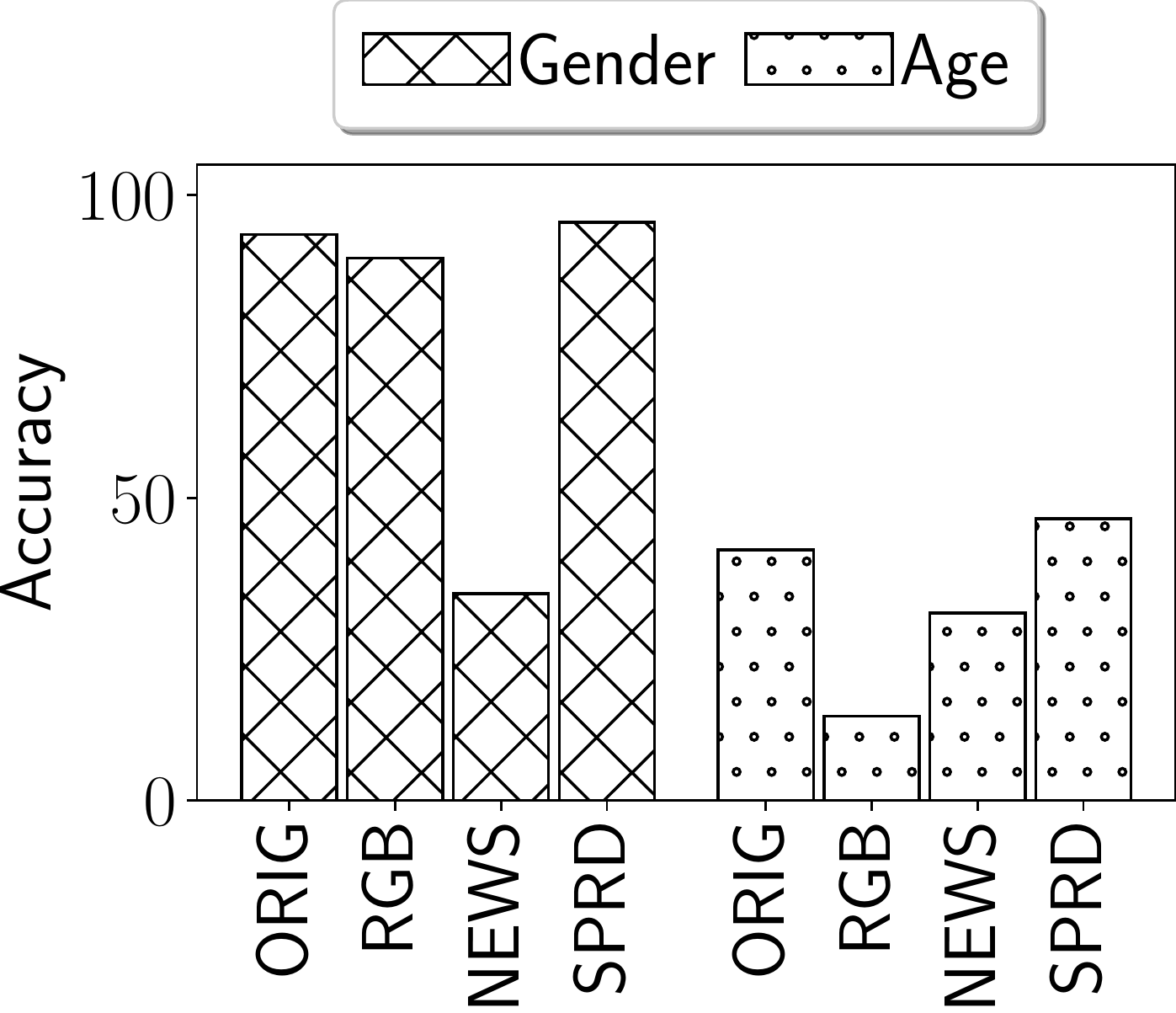}
		\caption{Accuracy on Face++}
		\label{Fig: FPP-cfd-accuracy}
	\end{subfigure}
	~\begin{subfigure}{0.66\columnwidth}
		\includegraphics[width= \textwidth, height=3.5cm, keepaspectratio]{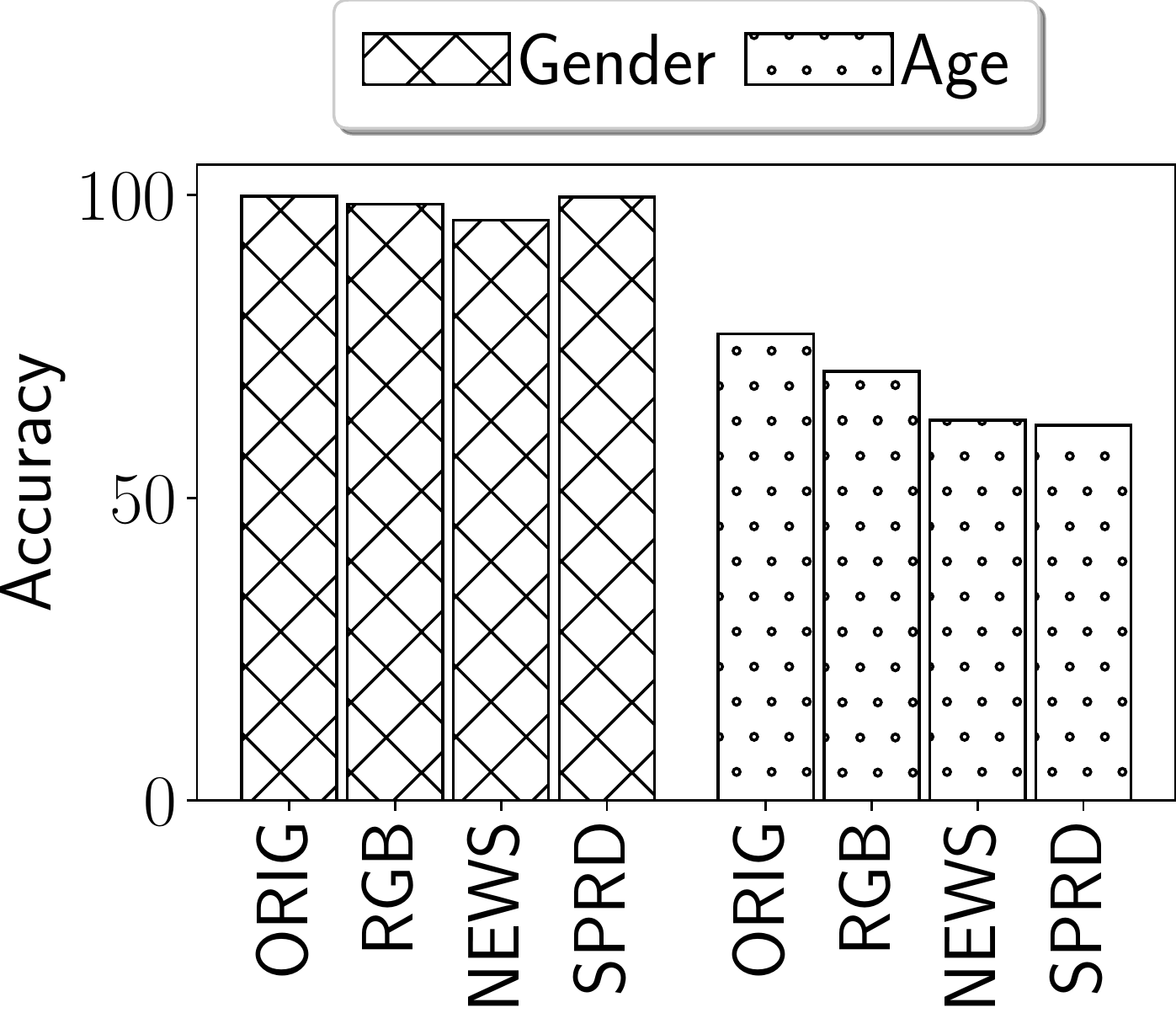}
		\caption{Accuracy on Microsoft}
		\label{Fig: MSFT-cfd-accuracy}
	\end{subfigure}
	
	\caption{{\bf Prediction accuracy on \textsc{CFD} for original, RGB-filter, newsprint-filter and spread-filter inputs. Under RGB and newsprint adversarial settings, all of the FRSs suffer from drop in accuracy (Face++ accuracy drops by $\bf > 50\%$), with Microsoft being marginally more robust than AWS and Face++.}}
	\label{fig:cfd-accuracy}
	\vspace{-4 mm}
\end{figure*}

The prediction accuracy results for the tasks of Gender and Age detection are presented in
Figure \ref{fig:cfd-accuracy}. The results are divided per FRS API in Figures \ref{Fig: AWS-cfd-accuracy}, \ref{Fig: FPP-cfd-accuracy} and \ref{Fig: MSFT-cfd-accuracy}.

\noindent
\textbf{Gender detection}: We observe that all the FRS APIs report high accuracy for gender detection. 
As regards to addition of different noises, we do not observe significant degradation in performance for RGB or spread noise. However, for the newsprint noise, we observe significant deterioration of prediction accuracy in Amazon and Face++ systems with a drop of $\geq 40\%$ (see Figures \ref{Fig: AWS-cfd-accuracy}, \ref{Fig: FPP-cfd-accuracy}). 

\noindent
\textbf{Age prediction}: Accuracy is low for all FRSs irrespective of the input. Here, Face++ is the worst performing system with the accuracy of nearly $14\%$ for Face++/RGB and nearly $ 47\%$ for Face++/spread (see Figure \ref{Fig: FPP-cfd-accuracy}). Microsoft, on the other hand, returns the best results with its reported accuracy in the range $\approx 62\%$ to $\approx 77\%$ (Figure \ref{Fig: MSFT-cfd-accuracy}).



\noindent
The key takeaways from Figure~\ref{fig:cfd-accuracy} are as follows.
\begin{compactitem}
	\item All APIs perform well for the task of gender detection for all inputs except newsprint, while the performance for age prediction is bad for all inputs on AWS and Face++.
	\item Face++ is the worst performing FRS for both gender and age prediction.
	\item Microsoft is the most robust FRS among the three.
\end{compactitem}


\if{0}
\begin{table*}[!ht]
	\small
	\begin{center}
		\begin{tabular}{c | c | c } 
			\hline
			\textbf{Image}\rule{0pt}{2ex} & \textbf{Gender} & \textbf{Age} \\
			\hline
			AWS/original\rule{0pt}{2ex} & \textbf{(W - B)} $3.48\%$& \textbf{(W - B)} $\textbf{23.64\%}$ \\ 
			AWS/RGB\rule{0pt}{2ex} & \textbf{(A - B)} $8.93\%$ & \textbf{(W - B)} $15.83\%$ \\
			AWS/newsprint\rule{0pt}{2ex} & \textbf{(A - B)} $\textbf{14.45\%}$ & \textbf{(A - B)} $22.66\%$ \\
			\cellcolor{red!20}Max. absolute increase in disparity\rule{0pt}{2ex} & \cellcolor{red!20} \textbf{10.97\% (newsprint)} & \cellcolor{red!20} $\phi$\\			
			\hline
			Face++/original\rule{0pt}{2ex} & \textbf{(A - B)} $8.93\%$& \textbf{(L - W)} $7.67\%$ \\ 
			Face++/RGB\rule{0pt}{2ex} & \textbf{(L - W)} $8.48\%$ & \textbf{(B - A)} $7.68\%$ \\
			Face++/newsprint\rule{0pt}{2ex} & \textbf{(L - B)} $\textbf{36.89\%}$ & \textbf{(L - B)} $\textbf{38.83\%}$ \\
			\cellcolor{red!20}Max. absolute increase in disparity\rule{0pt}{2ex} & \cellcolor{red!20} \textbf{27.96\% (newsprint)} & \cellcolor{red!20} \textbf{31.16\% (newsprint)}\\
			\hline
			Microsoft/original\rule{0pt}{2ex} &\textbf{(W/L/A - B)} $0.51\%$& \textbf{(A - B)} $\textbf{15.37\%}$\\ 
			Microsoft/RGB\rule{0pt}{2ex} & \textbf{(L/A - B)} $2.54\%$ & \textbf{(L - B)} $15.16\%$ \\
			Microsoft/Newsprint\rule{0pt}{2ex} & \textbf{(A - B)} $\textbf{4.26\%}$ & \textbf{(A - B)} $13.38\%$ \\
			\cellcolor{red!20}Max. absolute increase in disparity\rule{0pt}{2ex} & \cellcolor{red!20} \textbf{3.75\% (newsprint)} & \cellcolor{red!20} $\phi$\\
			\hline
		\end{tabular}
	\end{center}
	\caption{\label{tab:cfd-ethnic}\textbf{Difference in highest accuracy and lowest accuracy amongst the ethnic groups for \textsc{CFD}. Most results show significant disparity in prediction accuracy for people of `Black' ethnicity as compared to that of other ethnicities. Face++ reports the largest disparity for all tasks on the Newsprint adversarial input.}}
\end{table*}
\fi

\if{0}\subsubsection{Disparity in accuracy for ethnic groups}
\doubt{We are already having too many results and going over the desired page limits. If needed, I vote for dropping this section and Table~\ref{tab:cfd-ethnic}.}
We now evaluate the disparity in accuracy by taking the difference of the highest accuracy and the lowest accuracy among the different ethnic groups- `White', `Black', `Latino', `Asian'. The results are shown in Table~\ref{tab:cfd-ethnic}. 
Each cell in the table denote the highest disparity for a given task for the corresponding dataset. The mentions within parenthesis denote the ethnic groups between which the mentioned disparity was observed. Rows marked in red denote the maximum absolute increase in disparity from original to any of the two perturbed images.

\noindent
\textbf{Gender detection: } Apart from Face++ ($8.93\%$) both the other FRS do \textbf{not} report any significant disparity between accuracy for different groups. 
However, AWS and Face++ report extremely high disparities with the Newsprint filter-- $14.45\%$ and $ 36.89\%$ respectively. This is in contrast to Microsoft which has a disparity of only $\approx 4\%$. Note that much like the majority instances in \textsc{CelebSET}, here also we see that all the FRSs report this disparity against the `Black' ethnic group. 

\noindent
\textbf{Age prediction: }For age prediction task, both Amazon ($23.64\%$) and Microsoft ($15.37\%$) systems show significant disparity in accuracy for ethnic groups for the original images. Upon applying different noises (i.e., in adversarial set up), the disparity remains comparable throughout. However, for Face++ even though orignially the disparity was of $7.67\%$, upon exposing to newsprint filter, the disparity increases to almost 31\%. 
Again, much like in the prior cases, the disparities mostly exist against the `Black' ethnic group.

We observe similar results for biases between genders, the results for which have been omitted for brevity.

\noindent
\textbf{Summary: }
\begin{compactitem}
	\item All FRSs exhibit an increase in disparity of accuracy for Gender detection in adversarial set up. Disparity in case of age prediction remains comparable or increases (Face++) for the FRS.
	\item There is a noticeable and significant disparity in accuracy against individuals of `Black' ethnicity.
	\item Face++ reports the least accuracy as well as the highest inter-ethnic disparity. AWS and Microsoft exhibit all disparities against the `Black' ethnic group.
\end{compactitem}\fi


%
%

\begin{table*}[!ht]
	\small
	\begin{center}
		\begin{tabular}{c | c | c } 
			\hline
			\textbf{Image}\rule{0pt}{2ex} & \textbf{Gender} & \textbf{Age} \\
			\hline
			AWS/original\rule{0pt}{2ex} & \textbf{(WF/LM/AM - BF)} $7.69\%$& \textbf{(WM - BM)} $33.33\%$ \\ 
			AWS/RGB\rule{0pt}{2ex} & \textbf{(BM - BF)} $20.07\%$ & \textbf{(WF - AF)} $24.16\%$ \\
			AWS/newsprint\rule{0pt}{2ex} & \textbf{(WF - BM)} $\textbf{71.82\%}$ & \textbf{(LF - LM)} $\textbf{42.72\%}$ \\
			\cellcolor{red!20}Max. absolute increase in disparity\rule{0pt}{2ex} & \cellcolor{red!20} \textbf{64.13\% (newsprint)} & \cellcolor{red!20} \textbf{9.39\% (newsprint)}\\		
			\hline
			Face++/original\rule{0pt}{2ex} & \textbf{(WM/LM/AM - BF)} $21.25\%$ & \textbf{(LM - WF)} $28.80\%$ \\ 
			Face++/RGB\rule{0pt}{2ex} & \textbf{(WF/LF/AF - WM)} $25.81\%$ & \textbf{(BM - AF)} $21.91\%$ \\
			Face++/newsprint\rule{0pt}{2ex} & \textbf{(LM - BF)} $\textbf{93.27\%}$ & \textbf{(LF - BM)} $\textbf{44.26\%}$ \\
			\cellcolor{red!20}Max. absolute increase in disparity\rule{0pt}{2ex} & \cellcolor{red!20} \textbf{72.02\% (newsprint)} & \cellcolor{red!20} \textbf{15.46\% (newsprint)}\\
			\hline
			Microsoft/original\rule{0pt}{2ex} & \textbf{(WM/WF/BM/LM/LF/AM/AF - BF)} $0.96\%$& \textbf{(AF - BF)} $\textbf{25.22\%}$ \\ 
			Microsoft/RGB\rule{0pt}{2ex} & \textbf{(BM/LM/LF/AM/AF - BF)} $4.81\%$ & \textbf{(LF - BF)} $21.57\%$ \\
			Microsoft/newsprint\rule{0pt}{2ex} & \textbf{(BM/LF - BF)} $\textbf{11.54\%}$ & \textbf{(WM - BF)} $21.08\%$ \\
			\cellcolor{red!20}Max. absolute increase in disparity\rule{0pt}{2ex} & \cellcolor{red!20} \textbf{10.58\% (newsprint)} & \cellcolor{red!20} $\phi$\\
			\hline
		\end{tabular}
	\end{center}
	\caption{\label{tab:cfd-intersection}\textbf{Difference in highest accuracy and lowest accuracy among the intersectional groups for \textsc{CFD}. Most results show significant disparity in prediction accuracy for people of `Black' ethnicity (BF/BM) as compared to that of the other ethnic groups. Face++ reports the largest disparity for all tasks on the newsprint adversarial input.}}
	\vspace{- 5 mm}
\end{table*}

\subsubsection{Disparity in accuracy for intersectional groups}
Next, we discuss the disparity in accuracy among the different intersectional groups formed by taking male and female members from each of the ethnic groups -- `White', `Black', `Latino', `Asian'. The results are presented in Table \ref{tab:cfd-intersection}.
Each cell in the table denotes the highest disparity for a given task for the corresponding dataset. Once again, the mentions within parenthesis denote the ethnic groups between which the corresponding disparity was observed. Rows marked in red denote the maximum absolute increase in disparity from original to any of the two perturbed images. For paucity of space, we have reported the results for the newsprint and the RGB filters only.

\noindent
\textbf{Gender detection}: Barring Microsoft, both the other FRSs exhibit considerable disparity in accuracy in gender detection for original input images (21.25\% for Face++ against `Black' females). In the adversarial set up with RGB noise, the disparities increase considerably. However, upon usage of newsprint noise, the disparity among intersectional groups become scandalous. The maximum absolute increase in disparity goes well beyond 60\% for both Amazon and Face++. Much like our prior observations, `Black' males and Females are on the wrong side of the disparity. 


\noindent
\textbf{Age prediction}: For age prediction task, all the three systems show significant disparity in accuracy for ethnic groups for the original images ($\ge 25\%$ across FRSs). While all FRSs report a drop in disparity for the RGB filter, AWS and Face++ report an increase for newsprint, with Face++ reporting the highest disparity at $44\%$. Notice, even though there is a marginal drop in disparity, they are still comparable and more than 20\% among different intersectional groups. This on its own is very alarming. Microsoft FRSs always exhibits disparity against `Black' females; whereas AWS and Face++ show slightly different behaviour, with other minority groups too experiencing disparate accuracy.

\noindent
\textbf{Summary}
\begin{compactitem}
	\item All FRSs exhibit an increase in disparity of accuracy for gender detection, mostly directed toward individuals of `Black' ethnicity.
	\item Face++ reports the highest disparity for all experiments with a value of $93.27\%$ between `Latino' males and `Black' females.
	\item Microsoft is the best performing FRS, yet it is the most disparate against `Black' females, with all its lowest accuracy values being reported for this group.
\end{compactitem}


%
%

\subsection{Observations for \textsc{CFD-MR} and \textsc{CFD-India}}
Along with \textsc{CFD}, we also evaluate two other datasets -- \textsc{CFD-MR} and \textsc{CFD-India}. These are newly released datasets and allow us to study disparities in accuracy for ethnic groups that are not generally studied in the context of FRS audits. 

\subsubsection{Disparity in accuracy between males and females}

\begin{table*}
	\small
	\centering
	\begin{tabular}{c||c|c||c|c}
		\hline
		\textbf{Image} & \multicolumn{2}{c||}{\textsc{CFD-MR}}  & \multicolumn{2}{c}{\textsc{CFD-India}} \\
		\hline
		 & \textbf{Gender} & \textbf{Age}  & \textbf{Gender} & \textbf{Age} \\
		\hline
		AWS/original\rule{0pt}{2ex} & $3.23\%$ & $-10.04\%$ & $3.85\%$ & $15.77\%$\\ 
		AWS/RGB\rule{0pt}{2ex} &  $-2.24\%$ & $7.69\%$ & $7.69\%$ & $\textbf{23.08\%}$\\
		AWS/newsprint\rule{0pt}{2ex} &  $\textbf{-63.65\%}$ &  $\textbf{-11.91\%}$ &  $\textbf{-58.16\%}$ &  $-16.75\%$\\
		\cellcolor{red!20}Max. abs. inc. in disparity\rule{0pt}{2ex} & \cellcolor{red!20} \textbf{66.88\% (newsprint)} & \cellcolor{red!20} \textbf{17.73\% (RGB)} & \cellcolor{red!20} \textbf{62.01\% (newsprint)} & \cellcolor{red!20} \textbf{32.52\% (newsprint)}\\
		\hline
		Face++/original\rule{0pt}{2ex} & $12.90\%$ & $4.84\%$ & $19.23\%$ & $4.36\%$ \\ 
		Face++/RGB\rule{0pt}{2ex} & $-19.85\%$ & $8.93\%$ & $-0.30\%$ & $\textbf{-9.23\%}$\\
		Face++/newsprint\rule{0pt}{2ex} &  $\textbf{69.23\%}$ & $\textbf{29.03\%}$ &  $\textbf{62.90\%}$ & $-1.07\%$  \\
		\cellcolor{red!20}Max. abs. inc in disparity\rule{0pt}{2ex} & \cellcolor{red!20} \textbf{56.33\% (newsprint)} & \cellcolor{red!20} \textbf{24.19\% (newsprint)} & \cellcolor{red!20} \textbf{43.67\% (newsprint)} & \cellcolor{red!20} \textbf{13.59\%(RGB)}\\
		\hline
		Microsoft/original\rule{0pt}{2ex} & $0\%$&  $\textbf{18.12\%}$ & $0\%$&  $-0.64\%$\\ 
		Microsoft/RGB\rule{0pt}{2ex} &  $0\%$ &  $17.49\%$ &  $0\%$ &  $5.56\%$\\
		Microsoft/newsprint\rule{0pt}{2ex} &  $0\%$ &  $-7.2\%$ &  $\textbf{1.92\%}$ & $\textbf{17.48\%}$\\
		\cellcolor{red!20}Max. abs. inc. in disparity\rule{0pt}{2ex} & \cellcolor{red!20} \textbf{$\phi$} & \cellcolor{red!20} \textbf{25.32\% (newsprint)}& \cellcolor{red!20} \textbf{1.92\% (newsprint)} & \cellcolor{red!20} \textbf{18.12\% (newsprint)}\\
		\hline
	\end{tabular}
	\caption{\textbf{Difference in highest accuracy and lowest accuracy between males and females for \textsc{CFD-MR} and \textsc{CFD-India}. Most results show significant disparity in prediction accuracy toward males. AWS and Face++ report the largest disparities for all tasks on the newsprint adversarial input for both datasets and Microsoft reports no disparity in gender detection for \textsc{CFD-MR}.}} 
	\label{tab:cfd-mr-ind-intersection}
	\vspace*{-5mm}
\end{table*}

For brevity, we only discuss the disparities in accuracy between males and females in each of the datasets. The results for \textsc{CFD-MR} and \textsc{CFD-India} are shown in Table~\ref{tab:cfd-mr-ind-intersection}. 


\noindent
\textbf{Gender detection}: While Face++ has a disparity of $\ge 10\%$, AWS displays a marginal disparity ($\le 4\%$), and Microsoft system does not exhibit any disparity in accuracy for prediction of the original images for both \textsc{CFD-MR} and \textsc{CFD-India} datasets. However, upon imposing adversarial set up we see interesting and different behaviors across the different FRSs. While Microsoft is gracefully robust against all noises, AWS and Face++ are very sensitive in this regard (esp. for newsprint noise). While newsprint noise increases significant disparity ($-63.65\%$, $-58.16$ for \textsc{CFD-MR} and \textsc{CFD-India respectively}) in AWS system against photographs of males, the trend is completely reversed (significant discrimination against female photographs) in Face++.

\noindent
\textbf{Age prediction}: For \textsc{CFD-MR}, while AWS shows disparity in accuracy against males ($-10\%$), Microsoft shows disparity against females ($18\%$). However, apart from AWS system none of the other systems show considerable disparity on original images for \textsc{CFD-India} dataset. For RGB noice, all the systems show increasingly disparate performance toward females (except Face++ in \textsc{CFD-India}). The trend is completely reversed in newsprint noise set up, where majority of the disparity is against males (aberration Microsoft for \textsc{CFD-India}). However, regardless of the direction, in the adversarial set up we observe significant increase in disparity of accuracy and unlike gender detection task, here Microsoft system is the most sensitive to adversarial noises.

\noindent
\textbf{Summary:}
\begin{compactitem}
	\item We observe disparity in accuracy between genders for both the extension datasets - \textsc{CFD-MR} and \textsc{CFD-India}.
	\item Face++ is the most disparate FRS in \textsc{CFD-MR} while AWS is the most disparate in \textsc{CFD-India}.
	\item Newsprint filter results in the most instances of disparity.
\end{compactitem}

\section{Conclusion}
\label{sec:conclusion}
Our audit on three Facial Recognition Systems viz. Amazon AWS Rekognition, Face++ Detect and Microsoft Azure Face, using the \textsc{CelebSET}, \blu{ \textsc{FairFace}} and \textsc{CFD} datasets, shows that even with high prediction accuracy values, there exists a strong discriminatory bias against individuals of minority groups, specifically those belonging to `Black' ethnicity, for the tasks of gender, age and smile prediction. 
We also observe that for adversarial inputs generated from the same datasets, not only are the prediction accuracy values reduced significantly but the disparity in these values between different gender, ethnic or intersectional groups \textit{increase} by large margins. It should be noted that here again, the `Black' ethnic group is at the receiving end of this disparity more often than not.

\noindent
\textbf{Societal consequences}:
Our study has shown that the FRSs continue to display disparate accuracies despite repeated audits and while the accuracy of some tasks have improved, others have regressed, probably as a side effect. 
To this end, \textit{model cards}~\cite{mitchell2019model} have become an urgent necessity with the release of these FRSs so that they could be used in appropriate contexts only. 
 We believe that this is one of the best way forward in establishing trust with users and stress that it should be made mandatory for all future systems that have real life implications. The foremost goal of every FRS should be to have a fair behaviour and all efforts must be made in this direction while adhering to privacy and security needs of end users.
It is also imperative that organizations release white-papers or hold workshops explaining their technology so that the source of biases can be identified, and hopefully, mitigated.

\noindent
\textbf{Future directions}: An immediate next step in this work is to identify the correlation between strength of noise with change in bias by testing for different values of the parameters mentioned in the experiments section. \blu{Another future step would be to expand the scope of tasks to test for emotion prediction, face re-identification etc.,} and also study the impact of color filters to evaluate the sensitivity of FRSs to skin color in the images. 
Finally, we plan to evaluate with stronger adversarial generators which are specifically designed to fool such recognition systems.

\bibliography{paper.bib}

\begin{thebibliography}{59}
\providecommand{\natexlab}[1]{#1}
\providecommand{\url}[1]{\texttt{#1}}
\providecommand{\urlprefix}{URL }
\expandafter\ifx\csname urlstyle\endcsname\relax
  \providecommand{\doi}[1]{doi:\discretionary{}{}{}#1}\else
  \providecommand{\doi}{doi:\discretionary{}{}{}\begingroup
  \urlstyle{rm}\Url}\fi

\bibitem[{{Alexander Marrow}(2020)}]{marrow_frt2020}
{Alexander Marrow}. 2020.
\newblock ``Russia's lockdown surveillance measures need regulating, rights
  groups say'' Accessed April, 2021.
\newblock
  \urlprefix\url{https://www.reuters.com/article/us-health-coronavirus-russia-facial-reco-idUSKCN2253CM}.

\bibitem[{{Amazon}(2021)}]{aws_rekognition}
{Amazon}. 2021.
\newblock ``Amazon AWS Rekognition.'' Accessed April, 2021.
\newblock \urlprefix\url{https://aws.amazon.com/rekognition/faqs/}.

\bibitem[{Amazon(2021)}]{Amazon2021FAQ}
Amazon. 2021.
\newblock Amazon Rekognition FAQs.
\newblock https://aws.amazon.com/rekognition/faqs/.

\bibitem[{{Arwa Mahdawi}(2021)}]{arwa_guardian2021}
{Arwa Mahdawi}. 2021.
\newblock ``This AI-powered app will tell you if you're beautiful – and
  reinforce biases, too'' Accessed April, 2021.
\newblock
  \urlprefix\url{https://www.theguardian.com/commentisfree/2021/mar/06/ai-powered-app-tell-you-beautiful-reinforce-biases}.

\bibitem[{Barlas et~al.(2019)Barlas, Kyriakou, Kleanthous, and
  Otterbacher}]{BarlasICWSM2019}
Barlas, P.; Kyriakou, K.; Kleanthous, S.; and Otterbacher, J. 2019.
\newblock Social B(eye)as: Human and Machine Descriptions of People Images.
\newblock \emph{AAAI ICWSM} 13.

\bibitem[{Benjamin(2019)}]{benjamin2019race}
Benjamin, R. 2019.
\newblock Race after technology: Abolitionist tools for the new jim code.
\newblock \emph{Social Forces} .

\bibitem[{Bloice, Roth, and Holzinger(2019)}]{bloice2019augmentor}
Bloice, M.~D.; Roth, P.~M.; and Holzinger, A. 2019.
\newblock {Biomedical image augmentation using Augmentor}.
\newblock \emph{Bioinformatics} 4522--4524.

\bibitem[{Bolukbasi et~al.(2016)Bolukbasi, Chang, Zou, Saligrama, and
  Kalai}]{bolukbasi2016man}
Bolukbasi, T.; Chang, K.-W.; Zou, J.; Saligrama, V.; and Kalai, A. 2016.
\newblock Man is to computer programmer as woman is to homemaker? debiasing
  word embeddings.
\newblock \emph{arXiv preprint arXiv:1607.06520} .

\bibitem[{Bose and Aarabi(2018)}]{bose2018adversarial}
Bose, A.~J.; and Aarabi, P. 2018.
\newblock Adversarial attacks on face detectors using neural net based
  constrained optimization.
\newblock In \emph{IEEE MMSP}.

\bibitem[{Buolamwini(2018)}]{Buolamwini2018What}
Buolamwini, J. 2018.
\newblock When the Robot Doesn't See Dark Skin.
\newblock
  https://www.nytimes.com/2018/06/21/opinion/facial-analysis-technology-bias.html.

\bibitem[{Buolamwini and Gebru(2018)}]{buolamwini2018gender}
Buolamwini, J.; and Gebru, T. 2018.
\newblock Gender shades: Intersectional accuracy disparities in commercial
  gender classification.
\newblock In \emph{PMLR FAT*}.

\bibitem[{Chandrasekaran et~al.(2020)Chandrasekaran, Gao, Tang, Fawaz, Jha, and
  Banerjee}]{chandrasekaran2020faceoff}
Chandrasekaran, V.; Gao, C.; Tang, B.; Fawaz, K.; Jha, S.; and Banerjee, S.
  2020.
\newblock Face-Off: Adversarial Face Obfuscation.

\bibitem[{Chen et~al.(2021)Chen, Xie, Pang, He, and Tian}]{chen2021appending}
Chen, Z.; Xie, L.; Pang, S.; He, Y.; and Tian, Q. 2021.
\newblock Appending adversarial frames for universal video attack.
\newblock In \emph{Proceedings of the IEEE/CVF Winter Conference on
  Applications of Computer Vision}.

\bibitem[{Dash et~al.(2021)Dash, Chakraborty, Ghosh, Mukherjee, and
  Gummadi}]{dash2021when}
Dash, A.; Chakraborty, A.; Ghosh, S.; Mukherjee, A.; and Gummadi, K.~P. 2021.
\newblock When the Umpire is also a Player: Bias in Private Label Product
  Recommendations on E-commerce Marketplaces.
\newblock In \emph{ACM FAccT}.

\bibitem[{Dash, Mukherjee, and Ghosh(2019)}]{dash2019network}
Dash, A.; Mukherjee, A.; and Ghosh, S. 2019.
\newblock A network-centric framework for auditing recommendation systems.
\newblock In \emph{IEEE INFOCOM}.

\bibitem[{Duan et~al.(2020)Duan, Ma, Wang, Bailey, Qin, and
  Yang}]{Duan_2020_CVPR}
Duan, R.; Ma, X.; Wang, Y.; Bailey, J.; Qin, A.~K.; and Yang, Y. 2020.
\newblock Adversarial Camouflage: Hiding Physical-World Attacks With Natural
  Styles.
\newblock In \emph{IEEE/CVF CVPR}.

\bibitem[{{equalAIs}(2021)}]{equalais_2021}
{equalAIs}. 2021.
\newblock ``equalAIs Empowering Humans by Subverting Machines'' Accessed April,
  2021.
\newblock \urlprefix\url{https://equalais.media.mit.edu/}.

\bibitem[{Face++(2021)}]{FPP2021What}
Face++. 2021.
\newblock Face Attributes.
\newblock https://www.faceplusplus.com/attributes/.

\bibitem[{{Face++}(2021)}]{facepp}
{Face++}. 2021.
\newblock ``Face++ Detect.'' Accessed April, 2021.
\newblock \urlprefix\url{https://www.faceplusplus.com/face-detection/}.

\bibitem[{Garofalo et~al.(2018)Garofalo, Rimmer, Preuveneers, Joosen
  et~al.}]{garofalo2018fishy}
Garofalo, G.; Rimmer, V.; Preuveneers, D.; Joosen, W.; et~al. 2018.
\newblock Fishy faces: Crafting adversarial images to poison face
  authentication.
\newblock In \emph{$\{$USENIX$\}$ ($\{$WOOT$\}$ 18)}.

\bibitem[{{GIMP}(2021)}]{gimp}
{GIMP}. 2021.
\newblock ``GIMP:GNU Image Manipulation Program''.
\newblock \urlprefix\url{https://www.gimp.org/about/}.

\bibitem[{Goel et~al.(2018)Goel, Singh, Agarwal, Vatsa, and
  Singh}]{goel2018smartbox}
Goel, A.; Singh, A.; Agarwal, A.; Vatsa, M.; and Singh, R. 2018.
\newblock Smartbox: Benchmarking adversarial detection and mitigation
  algorithms for face recognition.
\newblock In \emph{IEEE BTAS}.

\bibitem[{Goodfellow, Shlens, and Szegedy(2014)}]{goodfellow2014explaining}
Goodfellow, I.~J.; Shlens, J.; and Szegedy, C. 2014.
\newblock Explaining and harnessing adversarial examples.
\newblock \emph{arXiv preprint arXiv:1412.6572} .

\bibitem[{{Jay Peters}(2020)}]{peters_verge2020}
{Jay Peters}. 2020.
\newblock ``IBM will no longer offer, develop, or research facial recognition
  technology'' Accessed April, 2021.
\newblock
  \urlprefix\url{https://www.theverge.com/2020/6/8/21284683/ibm-no-longer-general-purpose-facial-recognition-analysis-software}.

\bibitem[{Jiang et~al.(2019)Jiang, Ma, Chen, Bailey, and
  Jiang}]{jiang2019black}
Jiang, L.; Ma, X.; Chen, S.; Bailey, J.; and Jiang, Y.-G. 2019.
\newblock Black-box adversarial attacks on video recognition models.
\newblock In \emph{Proceedings of the 27th ACM International Conference on
  Multimedia}.

\bibitem[{Jung et~al.(2018)Jung, An, Kwak, Salminen, and
  Jansen}]{JungICWSM2018}
Jung, S.; An, J.; Kwak, H.; Salminen, J.; and Jansen, B. 2018.
\newblock Assessing the Accuracy of Four Popular Face Recognition Tools for
  Inferring Gender, Age, and Race .

\bibitem[{Karkkainen and Joo(2021)}]{karkkainenfairface}
Karkkainen, K.; and Joo, J. 2021.
\newblock FairFace: Face Attribute Dataset for Balanced Race, Gender, and Age
  for Bias Measurement and Mitigation.
\newblock In \emph{IEEE/CVF WACV}.

\bibitem[{Kyriakou et~al.(2019{\natexlab{a}})Kyriakou, Barlas, Kleanthous, and
  Otterbacher}]{kyriakou2019}
Kyriakou, K.; Barlas, P.; Kleanthous, S.; and Otterbacher, J.
  2019{\natexlab{a}}.
\newblock Fairness in Proprietary Image Tagging Algorithms: A Cross-Platform
  Audit on People Images.
\newblock \emph{AAAI ICWSM} .

\bibitem[{Kyriakou et~al.(2019{\natexlab{b}})Kyriakou, Barlas, Kleanthous, and
  Otterbacher}]{KyriakouICWSM2019}
Kyriakou, K.; Barlas, P.; Kleanthous, S.; and Otterbacher, J.
  2019{\natexlab{b}}.
\newblock Fairness in Proprietary Image Tagging Algorithms: A Cross-Platform
  Audit on People Images .

\bibitem[{Lakshmi et~al.(2021)Lakshmi, Wittenbrink, Correll, and
  Ma}]{lakshmi2021india}
Lakshmi, A.; Wittenbrink, B.; Correll, J.; and Ma, D.~S. 2021.
\newblock The India Face Set: International and Cultural Boundaries Impact Face
  Impressions and Perceptions of Category Membership.
\newblock \emph{Frontiers in psychology} .

\bibitem[{Ma, Correll, and Wittenbrink(2015)}]{ma2015chicago}
Ma, D.~S.; Correll, J.; and Wittenbrink, B. 2015.
\newblock The Chicago face database: A free stimulus set of faces and norming
  data.
\newblock \emph{Behavior research methods} .

\bibitem[{Ma, Kantner, and Wittenbrink(2020)}]{ma2020chicago}
Ma, D.~S.; Kantner, J.; and Wittenbrink, B. 2020.
\newblock Chicago Face Database: Multiracial expansion.
\newblock \emph{Behavior Research Methods} .

\bibitem[{Maesumi et~al.(2021)Maesumi, Zhu, Wang, Chen, Wang, and
  Bajaj}]{maesumi2021learning}
Maesumi, A.; Zhu, M.; Wang, Y.; Chen, T.; Wang, Z.; and Bajaj, C. 2021.
\newblock Learning Transferable 3D Adversarial Cloaks for Deep Trained
  Detectors.
\newblock \emph{arXiv preprint arXiv:2104.11101} .

\bibitem[{{Manish Singh}(2020)}]{singh_frt2021}
{Manish Singh}. 2020.
\newblock ``India used facial recognition tech to identify 1,100 individuals at
  a recent riot'' Accessed April, 2021.
\newblock
  \urlprefix\url{https://techcrunch.com/2020/03/11/india-used-facial-recognition-tech-to-identify-1100-individuals-at-a-recent-riot/}.

\bibitem[{Massoli et~al.(2021)Massoli, Carrara, Amato, and
  Falchi}]{massoli2021detection}
Massoli, F.~V.; Carrara, F.; Amato, G.; and Falchi, F. 2021.
\newblock Detection of face recognition adversarial attacks.
\newblock \emph{Computer Vision and Image Understanding} 202: 103103.

\bibitem[{Mehrabi et~al.(2019)Mehrabi, Morstatter, Saxena, Lerman, and
  Galstyan}]{mehrabi2019survey}
Mehrabi, N.; Morstatter, F.; Saxena, N.; Lerman, K.; and Galstyan, A. 2019.
\newblock A survey on bias and fairness in machine learning.
\newblock \emph{arXiv preprint arXiv:1908.09635} .

\bibitem[{Messias, Vikatos, and Benevenuto(2017)}]{MessiasWI2017}
Messias, J.; Vikatos, P.; and Benevenuto, F. 2017.
\newblock White, Man, and Highly Followed: Gender and Race Inequalities in
  Twitter.
\newblock In \emph{Proceedings of the International Conference on Web
  Intelligence}.

\bibitem[{{Microsoft}(2021)}]{microsoft_face}
{Microsoft}. 2021.
\newblock ``Microsoft Azure Face.'' Accessed April, 2021.
\newblock
  \urlprefix\url{https://azure.microsoft.com/en-in/services/cognitive-services/face/}.

\bibitem[{Microsoft(2021)}]{MS2021What}
Microsoft. 2021.
\newblock What is the Azure Face service?
\newblock
  https://docs.microsoft.com/en-us/azure/cognitive-services/face/overview.

\bibitem[{Mitchell et~al.(2019)Mitchell, Wu, Zaldivar, Barnes, Vasserman,
  Hutchinson, Spitzer, Raji, and Gebru}]{mitchell2019model}
Mitchell, M.; Wu, S.; Zaldivar, A.; Barnes, P.; Vasserman, L.; Hutchinson, B.;
  Spitzer, E.; Raji, I.~D.; and Gebru, T. 2019.
\newblock Model cards for model reporting.
\newblock In \emph{ACM FAT*}.

\bibitem[{Nagpal et~al.(2019)Nagpal, Singh, Singh, and Vatsa}]{nagpal2019deep}
Nagpal, S.; Singh, M.; Singh, R.; and Vatsa, M. 2019.
\newblock Deep learning for face recognition: Pride or prejudiced?
\newblock \emph{arXiv preprint arXiv:1904.01219} .

\bibitem[{{Natasha Lomas}(2020)}]{lomas_frt2020}
{Natasha Lomas}. 2020.
\newblock ``London’s Met Police switches on live facial recognition, flying
  in face of human rights concerns'' Accessed April, 2021.
\newblock
  \urlprefix\url{https://techcrunch.com/2020/01/24/londons-met-police-switches-on-live-facial-recognition-flying-in-face-of-human-rights-concerns/}.

\bibitem[{Noble(2018)}]{noble2018algorithms}
Noble, S.~U. 2018.
\newblock \emph{Algorithms of oppression: How search engines reinforce racism}.
\newblock nyu Press.

\bibitem[{Oh, Fritz, and Schiele(2017)}]{oh2017adversarial}
Oh, S.~J.; Fritz, M.; and Schiele, B. 2017.
\newblock Adversarial image perturbation for privacy protection a game theory
  perspective.
\newblock In \emph{IEEE ICCV}.

\bibitem[{O'neil(2016)}]{o2016weapons}
O'neil, C. 2016.
\newblock \emph{Weapons of math destruction: How big data increases inequality
  and threatens democracy}.
\newblock Crown.

\bibitem[{Pang et~al.(2015)Pang, Baretto, Kautz, and Luo}]{PangBigData2015}
Pang, R.; Baretto, A.; Kautz, H.; and Luo, J. 2015.
\newblock Monitoring adolescent alcohol use via multimodal analysis in social
  multimedia.
\newblock In \emph{IEEE Big Data}.

\bibitem[{Qiu et~al.(2020)Qiu, Xiao, Yang, Yan, Lee, and
  Li}]{qiu2019semanticadv}
Qiu, H.; Xiao, C.; Yang, L.; Yan, X.; Lee, H.; and Li, B. 2020.
\newblock Semanticadv: Generating adversarial examples via
  attribute-conditioned image editing.
\newblock In \emph{ECCV}.

\bibitem[{Raji and Buolamwini(2019)}]{raji2019action}
Raji, I.~D.; and Buolamwini, J. 2019.
\newblock Actionable Auditing: Investigating the Impact of Publicly Naming
  Biased Performance Results of Commercial AI Products.
\newblock In \emph{AAAI/ACM AIES}.

\bibitem[{Raji et~al.(2020)Raji, Gebru, Mitchell, Buolamwini, Lee, and
  Denton}]{raji2020saving}
Raji, I.~D.; Gebru, T.; Mitchell, M.; Buolamwini, J.; Lee, J.; and Denton, E.
  2020.
\newblock Saving face: Investigating the ethical concerns of facial recognition
  auditing.
\newblock In \emph{AAAI/ACM AIES}.

\bibitem[{Sandvig et~al.(2014)Sandvig, Hamilton, Karahalios, and
  Langbort}]{sandvig2014auditing}
Sandvig, C.; Hamilton, K.; Karahalios, K.; and Langbort, C. 2014.
\newblock Auditing algorithms: Research methods for detecting discrimination on
  internet platforms.
\newblock \emph{Data and discrimination: converting critical concerns into
  productive inquiry} .

\bibitem[{Sangomla(2020)}]{Sangomla2020Big}
Sangomla, A. 2020.
\newblock Big Brother is watching you; actually your face.
\newblock
  https://www.downtoearth.org.in/news/science-technology/big-brother-is-watching-you-actually-your-face-74181.

\bibitem[{Sixta et~al.(2020)Sixta, Junior, Buch-Cardona, Vazquez, and
  Escalera}]{sixta2020fairface}
Sixta, T.; Junior, J. C.~J.; Buch-Cardona, P.; Vazquez, E.; and Escalera, S.
  2020.
\newblock Fairface challenge at ECCV 2020: analyzing bias in face recognition.
\newblock In \emph{ECCV}.

\bibitem[{Snow(2018)}]{Snow2018Amazon}
Snow, J. 2018.
\newblock Amazon’s Face Recognition Falsely Matched 28 Members of Congress
  With Mugshots.
\newblock
  https://www.aclu.org/blog/privacy-technology/surveillance-technologies/amazons-face-recognition-falsely-matched-28.

\bibitem[{{STOP}(2021)}]{stop_2021}
{STOP}. 2021.
\newblock ``Surveillance Technology Oversight Project'' Accessed April, 2021.
\newblock \urlprefix\url{https://www.stopspying.org/}.

\bibitem[{{Tate Ryan-Mosley}(2021)}]{tate_mit2021}
{Tate Ryan-Mosley}. 2021.
\newblock ``I asked an AI to tell me how beautiful I am'' Accessed April, 2021.
\newblock
  \urlprefix\url{https://www.technologyreview.com/2021/03/05/1020133/ai-algorithm-rate-beauty-score-attractive-face/}.

\bibitem[{Vakhshiteh, Nickabadi, and
  Ramachandra(2020)}]{vakhshiteh2020adversarial}
Vakhshiteh, F.; Nickabadi, A.; and Ramachandra, R. 2020.
\newblock Adversarial Attacks against Face Recognition: A Comprehensive Study.
\newblock \emph{arXiv preprint arXiv:2007.11709} .

\bibitem[{Vikatos et~al.(2017)Vikatos, Messias, Miranda, and
  Benevenuto}]{VikatosHT2017}
Vikatos, P.; Messias, J.; Miranda, M.; and Benevenuto, F. 2017.
\newblock Linguistic Diversities of Demographic Groups in Twitter.
\newblock In \emph{ACM HT}.

\bibitem[{Xiao et~al.(2019)Xiao, Yang, Li, Deng, and Liu}]{xiao2019meshadv}
Xiao, C.; Yang, D.; Li, B.; Deng, J.; and Liu, M. 2019.
\newblock Meshadv: Adversarial meshes for visual recognition.
\newblock In \emph{IEEE/CVF CVPR}.

\bibitem[{Xu et~al.(2020)Xu, Chen, Xiao, Gao, Shen, and Shen}]{Xu_2020_CVPR}
Xu, X.; Chen, J.; Xiao, J.; Gao, L.; Shen, F.; and Shen, H.~T. 2020.
\newblock What Machines See Is Not What They Get: Fooling Scene Text
  Recognition Models With Adversarial Text Images.
\newblock In \emph{IEEE/CVF CVPR}.

\end{thebibliography}

\section*{Appendix}

We evaluate the performance of the three commercial FRS on both \textsc{CFD-MR} and \textsc{CFD-India} datasets. The results are discussed in Figures~\ref{fig:cfd-mr-accuracy} and ~\ref{fig:cfd-ind-accuracy}.
\noindent
\subsubsection{Effect of perturbations in the performance}
\begin{figure*}[t]
	\centering
	\begin{subfigure}{0.66\columnwidth}
		\includegraphics[width= \textwidth, height=4.5cm, keepaspectratio]{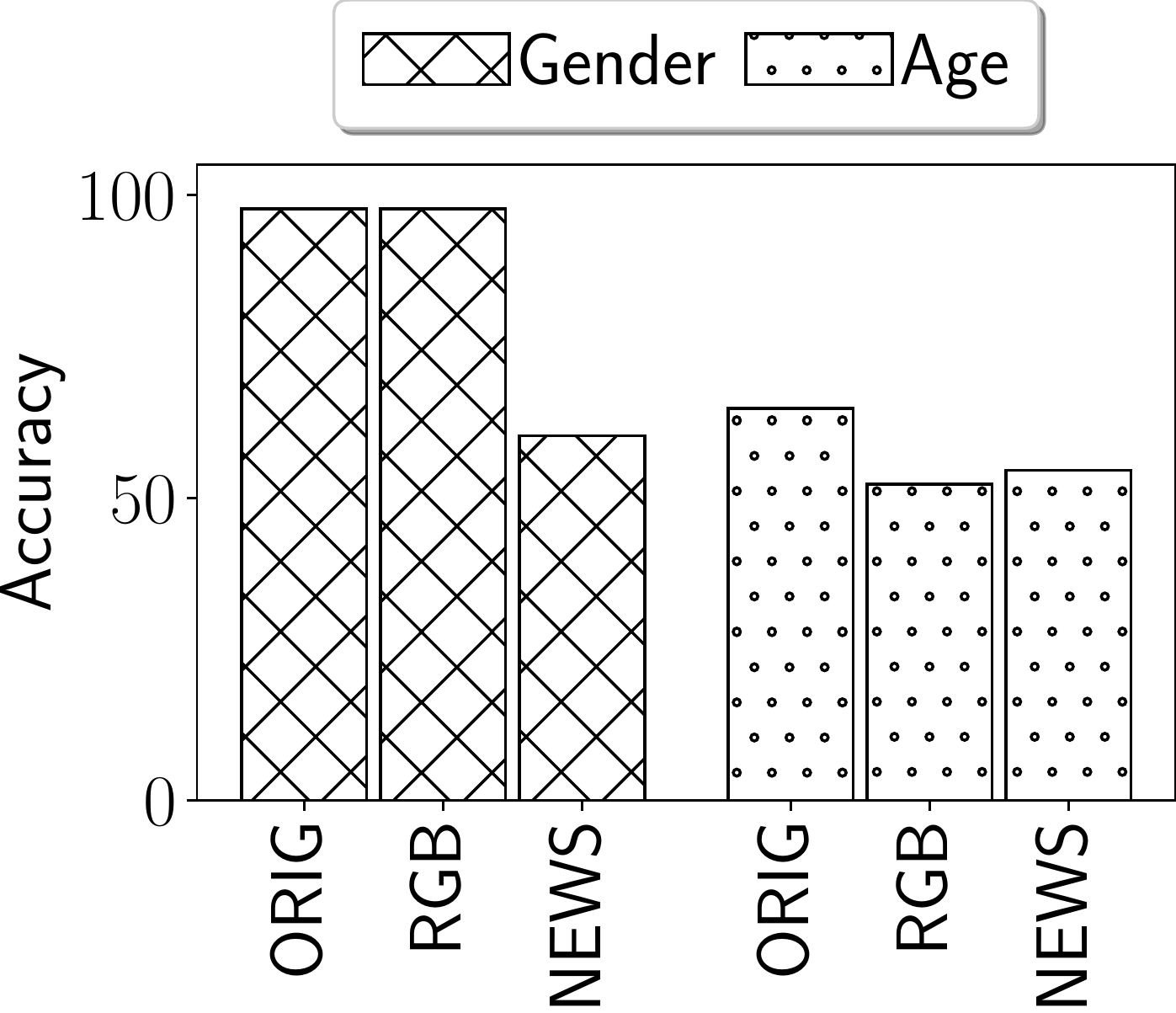}
		\caption{Accuracy on AWS}
		\label{Fig: AWS-cfd-mr-accuracy}
	\end{subfigure}%
	~\begin{subfigure}{0.66\columnwidth}
		\centering
		\includegraphics[width= \textwidth, height=4.5cm, keepaspectratio]{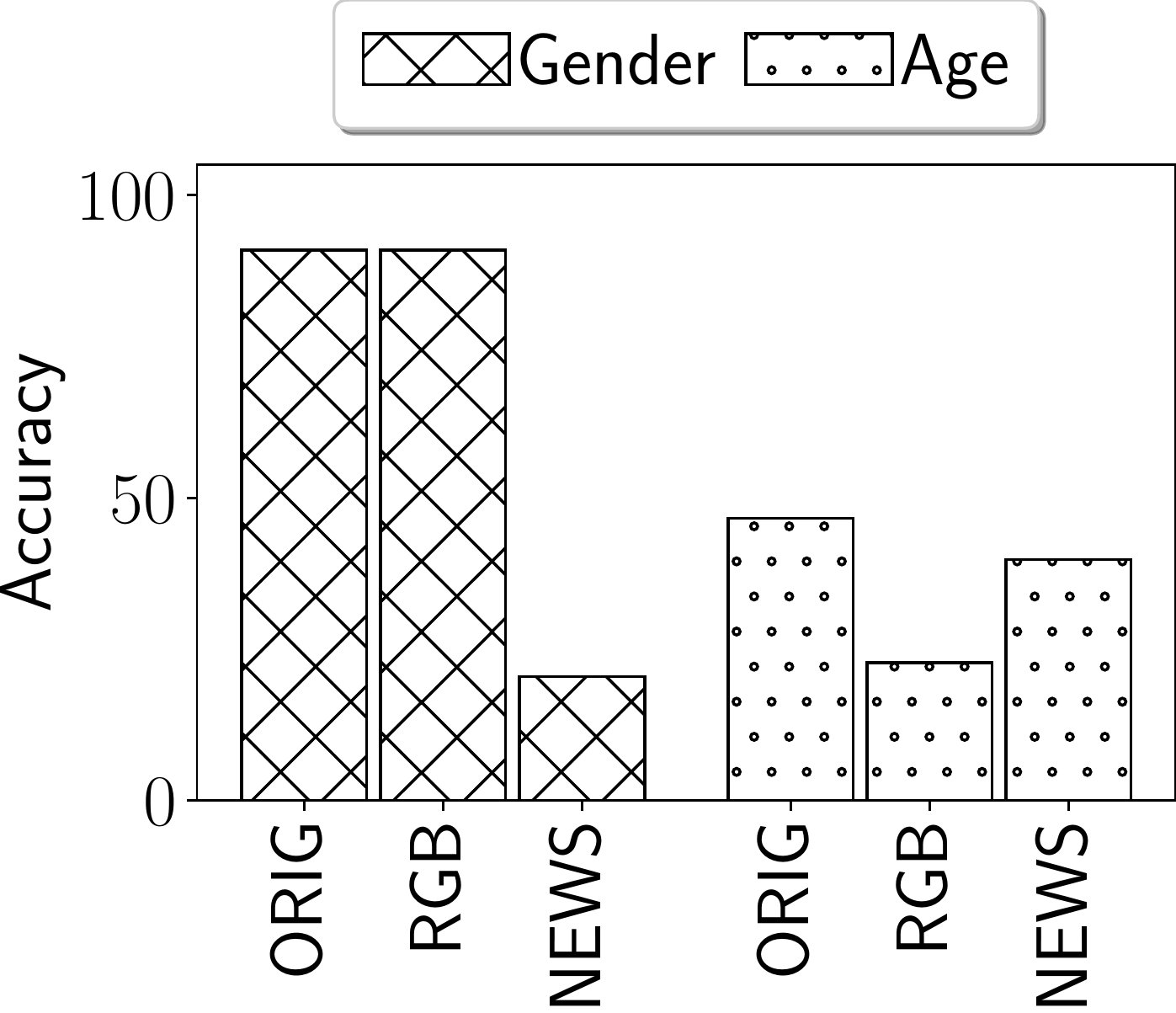}
		\caption{Accuracy on Face++}
		\label{Fig: FPP-cfd-mr-accuracy}
	\end{subfigure}
	~\begin{subfigure}{0.66\columnwidth}
		\centering
		\includegraphics[width= \textwidth, height=4.5cm, keepaspectratio]{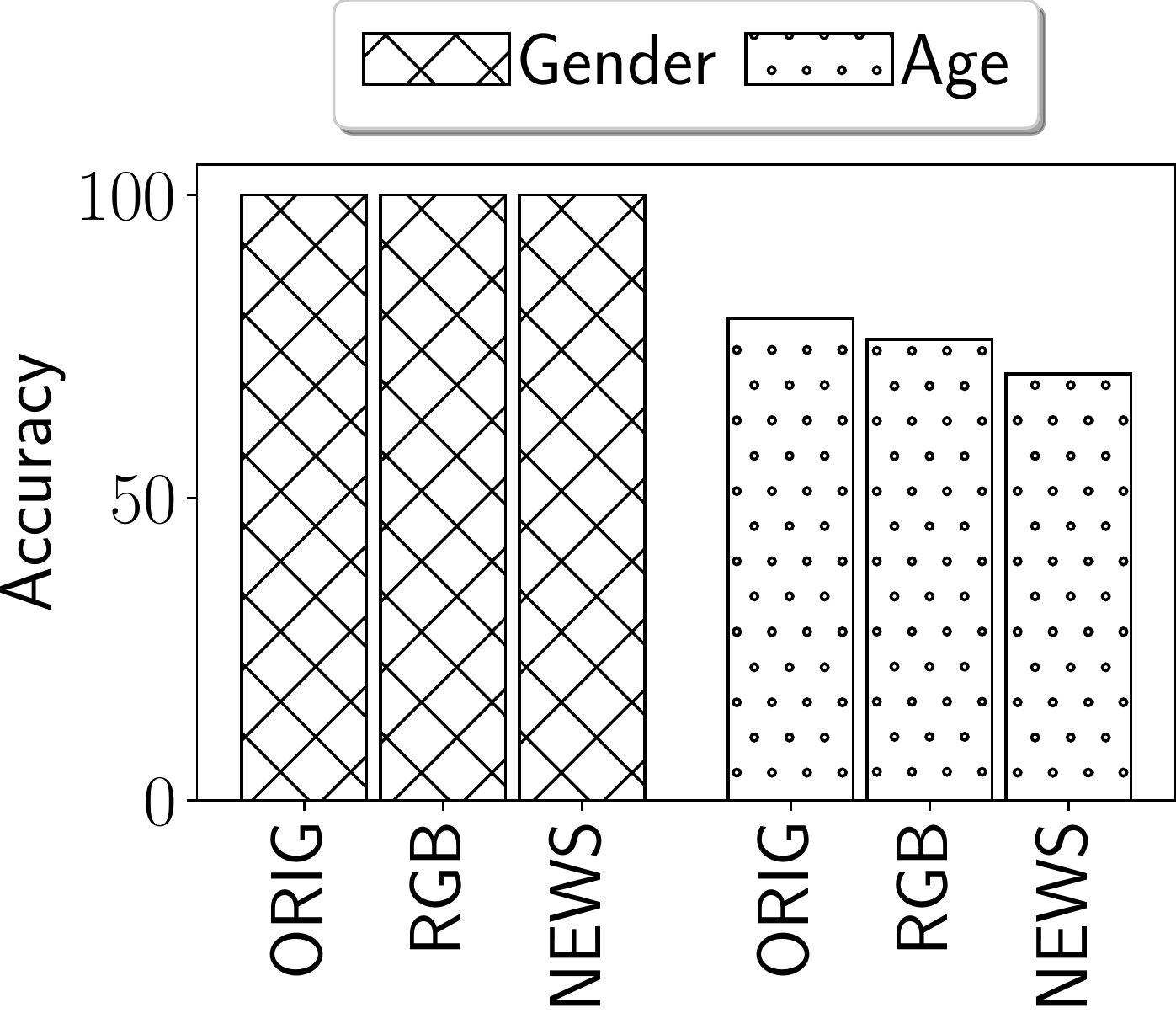}
		\caption{Accuracy on Microsoft}
		\label{Fig: MSFT-cfd-mr-accuracy}
	\end{subfigure}

	\caption{{\bf Performance on \textsc{CFD-MR} for Original, RGB-Filter and Newsprint-filter inputs }}
	\label{fig:cfd-mr-accuracy}
\end{figure*}

\begin{figure*}[t]
	\centering
	\begin{subfigure}{0.66\columnwidth}
		\includegraphics[width= \textwidth, height=4.5cm, keepaspectratio]{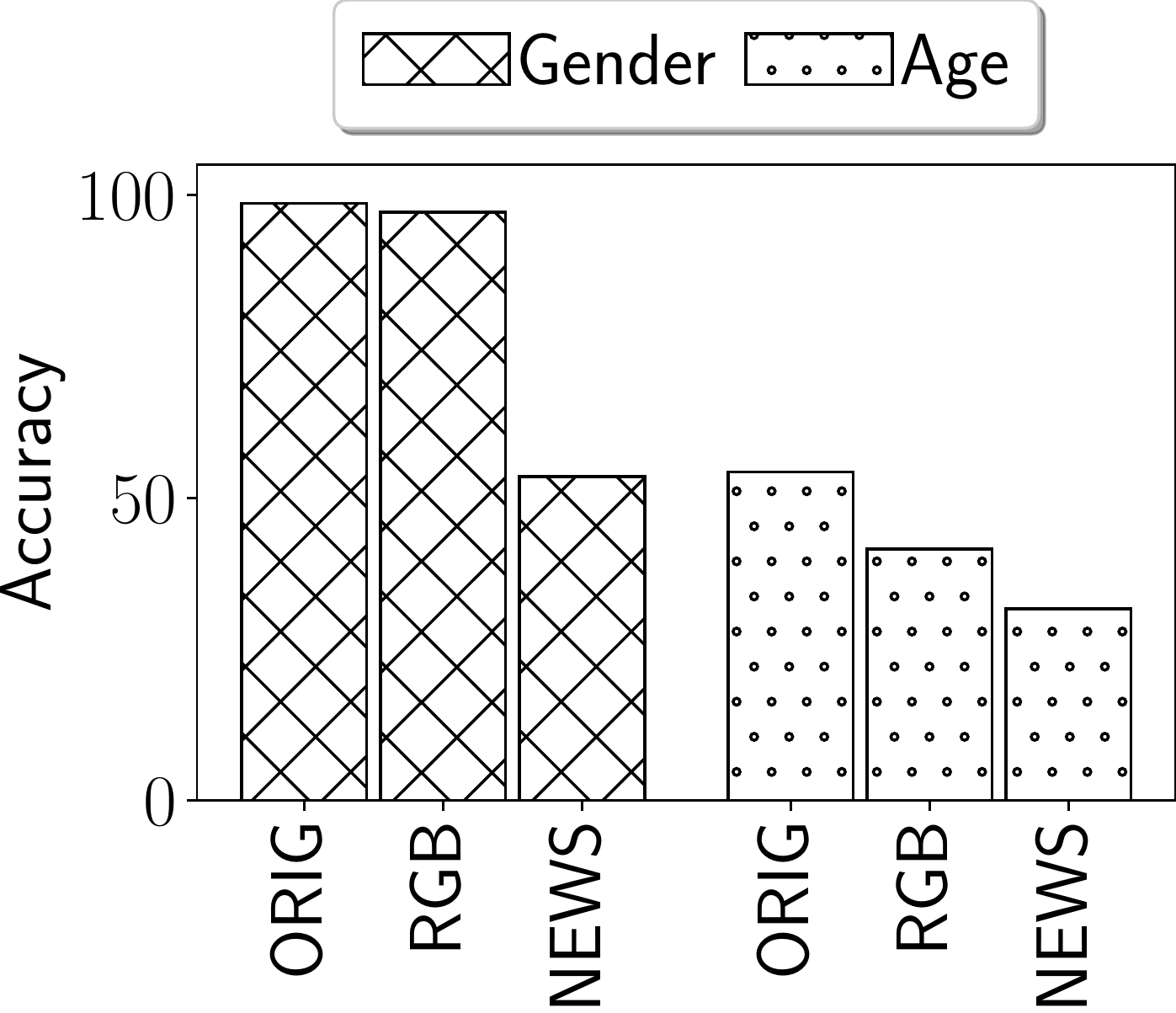}
		\caption{Accuracy on AWS}
		\label{Fig: AWS-cfd-ind-accuracy}
	\end{subfigure}%
	~\begin{subfigure}{0.66\columnwidth}
		\centering
		\includegraphics[width= \textwidth, height=4.5cm, keepaspectratio]{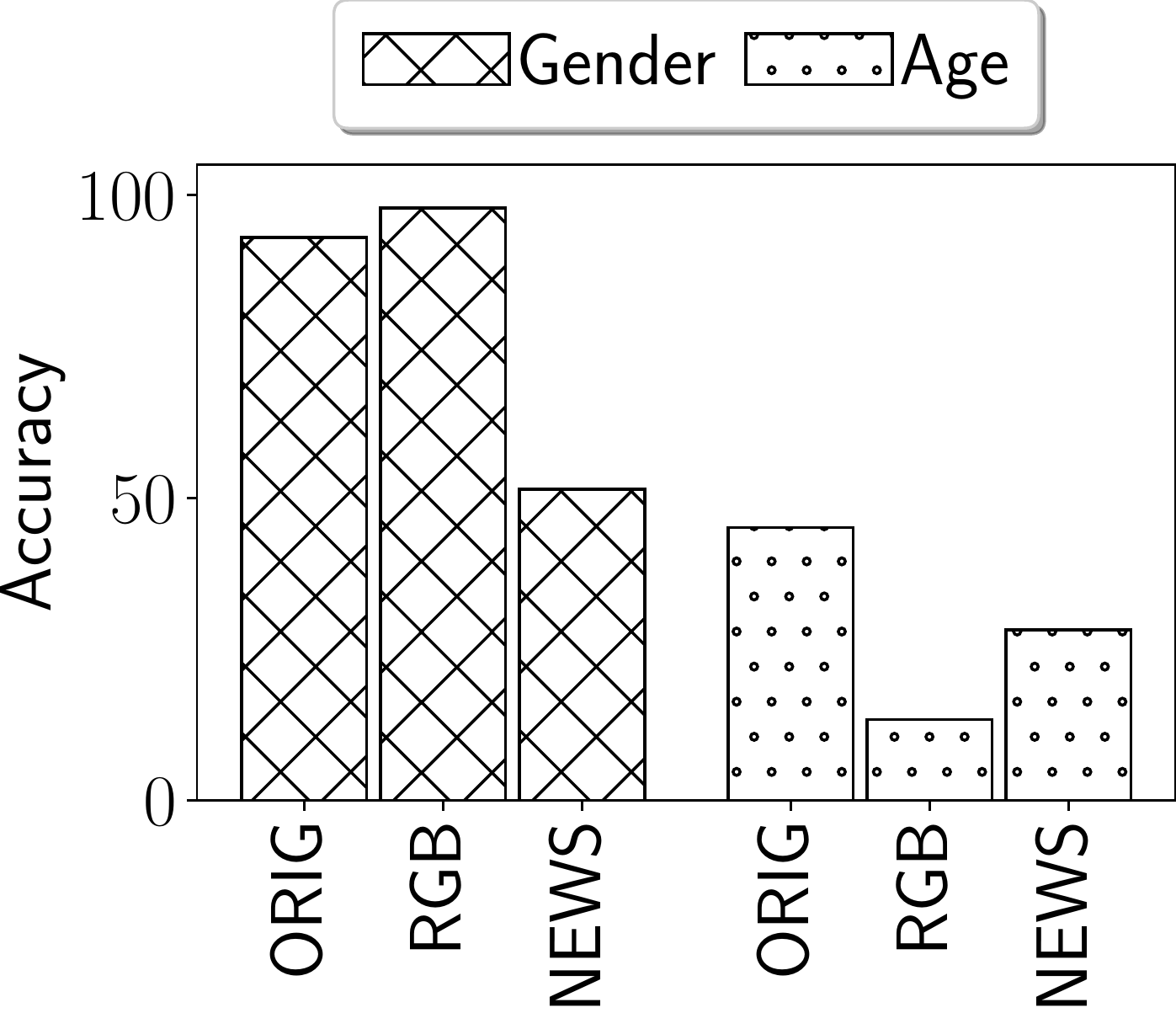}
		\caption{Accuracy on Face++}
		\label{Fig: FPP-cfd-ind-accuracy}
	\end{subfigure}
	~\begin{subfigure}{0.66\columnwidth}
		\centering
		\includegraphics[width= \textwidth, height=4.5cm, keepaspectratio]{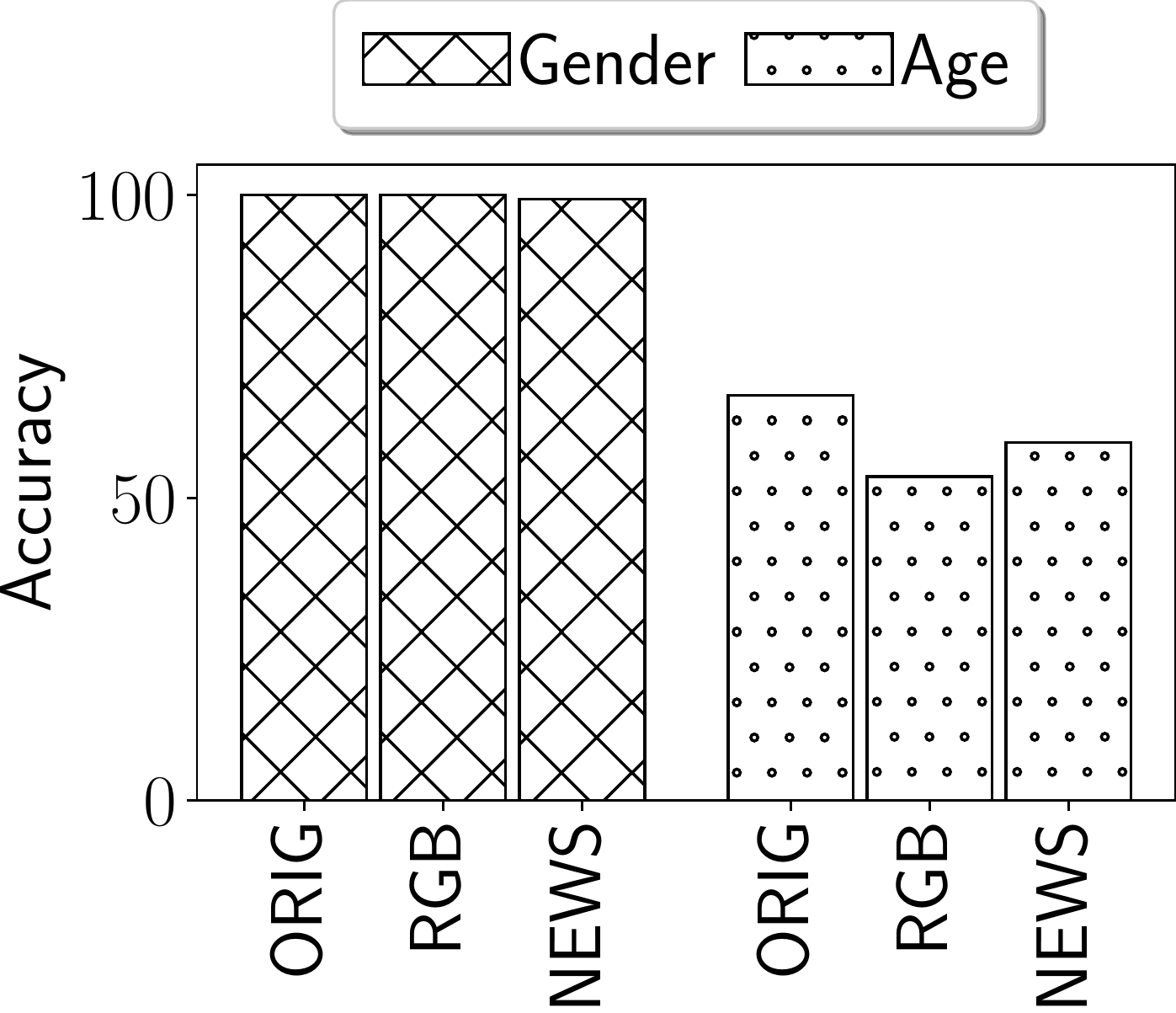}
		\caption{Accuracy on Microsoft}
		\label{Fig: MSFT-cfd-ind-accuracy}
	\end{subfigure}

	\caption{{\bf Performance on \textsc{CFD-India} for Original, RGB-Filter and Newsprint-filter inputs }}
	\label{fig:cfd-ind-accuracy}
\end{figure*}

The prediction accuracy results for the tasks of Gender and Age detection are presented here. The results are divided per FRS API in Figures \ref{Fig: AWS-cfd-mr-accuracy}, \ref{Fig: FPP-cfd-mr-accuracy} and \ref{Fig: MSFT-cfd-mr-accuracy} for \textsc{CFD-MR} and in Figures \ref{Fig: AWS-cfd-ind-accuracy}, \ref{Fig: FPP-cfd-ind-accuracy} and \ref{Fig: MSFT-cfd-ind-accuracy} for \textsc{CFD-India}.

\textbf{Gender detection}: We observe that all the FRS APIs report high accuracy for gender detection on the original set of images for both the datasets ($\geq 90\%$), with Face++ reporting comparatively lowest accuracy values-- $91\%$ for \textsc{CFD-MR} (Fig.\ref{Fig: FPP-cfd-mr-accuracy}) and $93\%$ for \textsc{CFD-India} (Fig.\ref{Fig: FPP-cfd-ind-accuracy}). 
As regards to addition of different noises, we do not observe significant degradation in performance for RGB noise, with the accuracy for \textsc{CFD-India} increasing by $5\%$. However, for the newsprint noise, we observe significant deterioration of prediction accuracy in Amazon and Face++ systems with a drop of $\geq 40\%$ and Face++ reporting an accuracy of $\approx 20\%$ for \textsc{CFD-MR} (see Figures \ref{Fig: AWS-cfd-mr-accuracy}, \ref{Fig: FPP-cfd-mr-accuracy}, \ref{Fig: AWS-cfd-ind-accuracy} and \ref{Fig: FPP-cfd-ind-accuracy}). 

\noindent
\textbf{Age prediction}: The accuracy for this task is low for all FRSs, independent of the input. Face++ is the worst performing system here, similar to its performance on \textsc{CFD}, with the accuracy nearly $15\%$ and $13\%$ for RGB noise on \textsc{CFD-MR} and \textsc{CFD-India} respectively (see Figure \ref{Fig: FPP-cfd-mr-accuracy} and \ref{Fig: FPP-cfd-ind-accuracy}). Here, Microsoft returns the highest accuracy with its reported accuracy in the range of $\approx 76\%$ to $\approx 80\%$ for \textsc{CFD-MR} (see Figure \ref{Fig: MSFT-cfd-mr-accuracy}) and $\approx 54\%$ to $\approx 67\%$ for \textsc{CFD-India} (see Figure \ref{Fig: MSFT-cfd-ind-accuracy})

\noindent
The key takeaways from Figures~\ref{fig:cfd-mr-accuracy} and~\ref{fig:cfd-ind-accuracy}  are as follows.
\begin{compactitem}
	\item All APIs perform well for the task of gender detection for all inputs except newsprint, while the performance for age prediction is bad for all inputs on AWS and Face++.
	\item Face++ is the worst performing FRS for both gender and age prediction.
	\item Microsoft is the most robust FRS among the three.
\end{compactitem}
\end{document}